\documentclass[12pt]{iopart}
\usepackage{cite}
\usepackage{longtable,float}
\usepackage{graphicx}
\usepackage{mathabx}
\usepackage{mathptmx} 
\usepackage{multirow}
\usepackage{color}
\usepackage{xcolor}
\usepackage[colorlinks, citecolor=black]{hyperref}
\usepackage{caption}
\usepackage{adjustbox}
\usepackage{upgreek}

\begin{document}

\title{Fish lateral line inspired perception and flow-aided control: A review}

\author{Yufan Zhai$^1$, Xingwen Zheng$^1$, Guangming Xie$^{1,2,3}$}
\address{$^1$ State Key Laboratory for Turbulence and Complex Systems, Intelligent Biomimetic Design Lab, College of Engineering, Peking University, Beijing, 100871, China}
\address{$^2$ Institute of Ocean Research, Peking University, Beijing, 100871, China.}
\address{$^3$ Peng Cheng Laboratory, 518055 Shenzhen, China.}

\ead{zhengxingwen@pku.edu.cn; xiegming@pku.edu.cn}

\vspace{10pt}
\begin{indented}
\item[]June 2020
\end{indented}

\begin{abstract}
Any phenomenon in nature is potential to be an inspiration for us to propose new ideas. Lateral line is a typical example which has attracted more interest in recent years. With the aid of lateral line, fish is capable of acquiring fluid information around, which is of great significance for them to survive, communicate and hunt underwater. In this paper, we briefly introduce the morphology and mechanism of the lateral line first. Then we focus on the development of artificial lateral line which typically consists of an array of sensors and can be installed on underwater robots. A series of sensors inspired by the lateral line with different sensing principles have been summarized. And then the applications of artificial lateral line system in hydrodynamic environment sensing and vortices detection, dipole oscillation source detection, and autonomous control of underwater robots have been surveyed. In addition, the existing problems and future foci in the field have been further discussed in detail. The current works and future foci have demonstrated that artificial lateral line has great potentials of research and contributes to the applications of underwater robots.
\end{abstract}

\section{Introduction}

Comparing with extracting land resources, it's more difficult to exploit marine resources. Due to the complexity of underwater environment, the causticity of the seawater, the high pressure of the deep seafloor, the poor visibility in the sea, and the strong interference to the sensors, people are facing extremely harsh conditions when exploiting marine resources. With the rapid development of machine manufacturing and artificial intelligence, robots serve significant functions in resource exploitation.

\begin{figure}[htbp]
	\centering
	\includegraphics[width=\columnwidth]{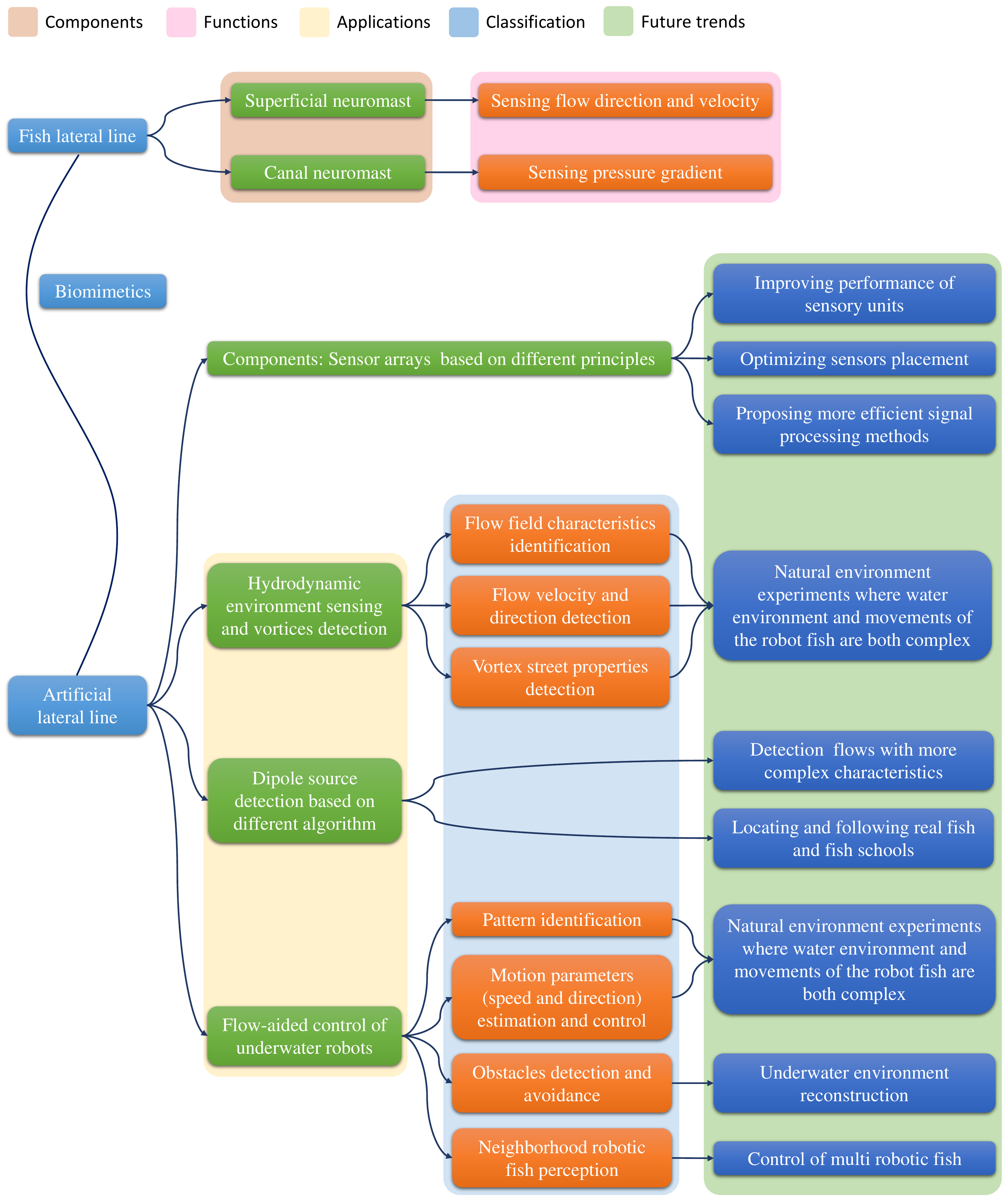}
	\caption{The graphical abstract of this article.}
	\label{Fig 0}
\end{figure}

However, the development of underwater robots is more difficult than that of land robots because of the above-mentioned reasons. With the further development of marine resources, the disadvantages of traditional underwater robots are gradually highlighted: large size, complex system, low control flexibility, high energy consumption, low efficiency, and lack of autonomy.

Biomimetics which acts as a new comprehensive subject provides another way for underwater robot manufacturing. Many researchers have taken inspiration from aquatic organisms and developed new underwater robots from the bionic perspective, commonly known as biomimetic underwater robots. For aquatic organisms, they have lived in marine environment for billions of years and adapted to the environment through evolution. They have the advantages of high maneuverability, high sensitivity to changes in surrounding environment and high energy efficiency, etc. Inspired by the excellent performance mentioned above, various kinds of biomimetic underwater robots have been developed successfully: robotic fish \cite{95}\cite{96}\cite{97}\cite{98}, robotic dolphin \cite{99}, robotic snake \cite{100}, robotic turtles \cite{101}, robotic salamanders \cite{102}, etc.

Owing to the complexity of underwater environment, robots must be equipped with well-developed underwater sensing system, which can help them to adapt to the environment and perform underwater tasks. Due to the particularity of water environment, land sensors cannot be directly used under the water, which restricts the development of sensing technology for underwater robots to some extent. To solve this problem, scientists have taken inspiration from marine life once again. Over billions of years of evolution in the ocean, fish have developed advanced systems for sensing the water environment. Lateral line is such a unique sensing system of fish and plays a key role for fish behaviours in complex water environment. The research and development of artificial lateral line (ALL) system referring to an array of sensors installed on the underwater vehicle will greatly improve the perception ability of the robots, which is advantageous for underwater missions.

In this paper, we firstly describe the biology of lateral line and the models which have been established to illustrate the mechanisms in section 2. In section 3, we review the sensors that are inspired by the lateral line and based on different sensing principles according to the categories. In section 4, we present applications of ALL in hydrodynamic environment sensing and vortices detection. In section 5, we introduce results in dipole oscillation source detection with the help of ALL using different methods. Finally, we discuss the flow-aided control of underwater robots using ALL system for different purposes. At the end of the article, we discuss the problems of current studies and try to put forward possible improvement strategies for further studies. The structure of the article is shown is Figure \ref{Fig 0}. The abbreviations used in this article and corresponding full names are listed in Table 1.

\begin{center}
	\begin{longtable}{p{8em}p{28em}}
		\caption{The abbreviations used in this article and corresponding full names}\\
		\hline
		Abbreviation &Full name \\
		\hline
		AlN	& Aluminium Nitride\\
		AHC & Artificial hair cell\\
		ALL & Artificial lateral line\\
		CN & Canal neuromast\\
		CRB	& Cramer-Rao bound\\
		CRLB & Cramer-Rao lower bound\\
		HWA	& Hot-wire anemometer\\
		IMU	& Inertial measurement unit\\
		IPMC & Ionic polymer-metal composite\\
		LCP & Liquid crystal polymer\\
		MEMS & Micro-electromechanical systems\\
		MMSE & Minimum mean-squared error\\
		MVDR & Minimum variance distortionless response\\
		MUSIC & Multiple signal classification\\
		PDMA & Plastic deformation magnetic assembly\\
		PDMS & Polydimethylsiloxane\\
		PVC & Polyvinyl chloride\\
		PVDF & Polyvinylidene fluoride\\
		SOI & Silicon-On-Insulator\\
		SN & Superficial neuromast\\
		SMD & Surface mounted devices\\
		\hline
	\end{longtable}
\end{center}

\section{Mechanisms and models of the fish lateral line}

\begin{figure}[htbp]
	\centering
	\includegraphics[width=\columnwidth]{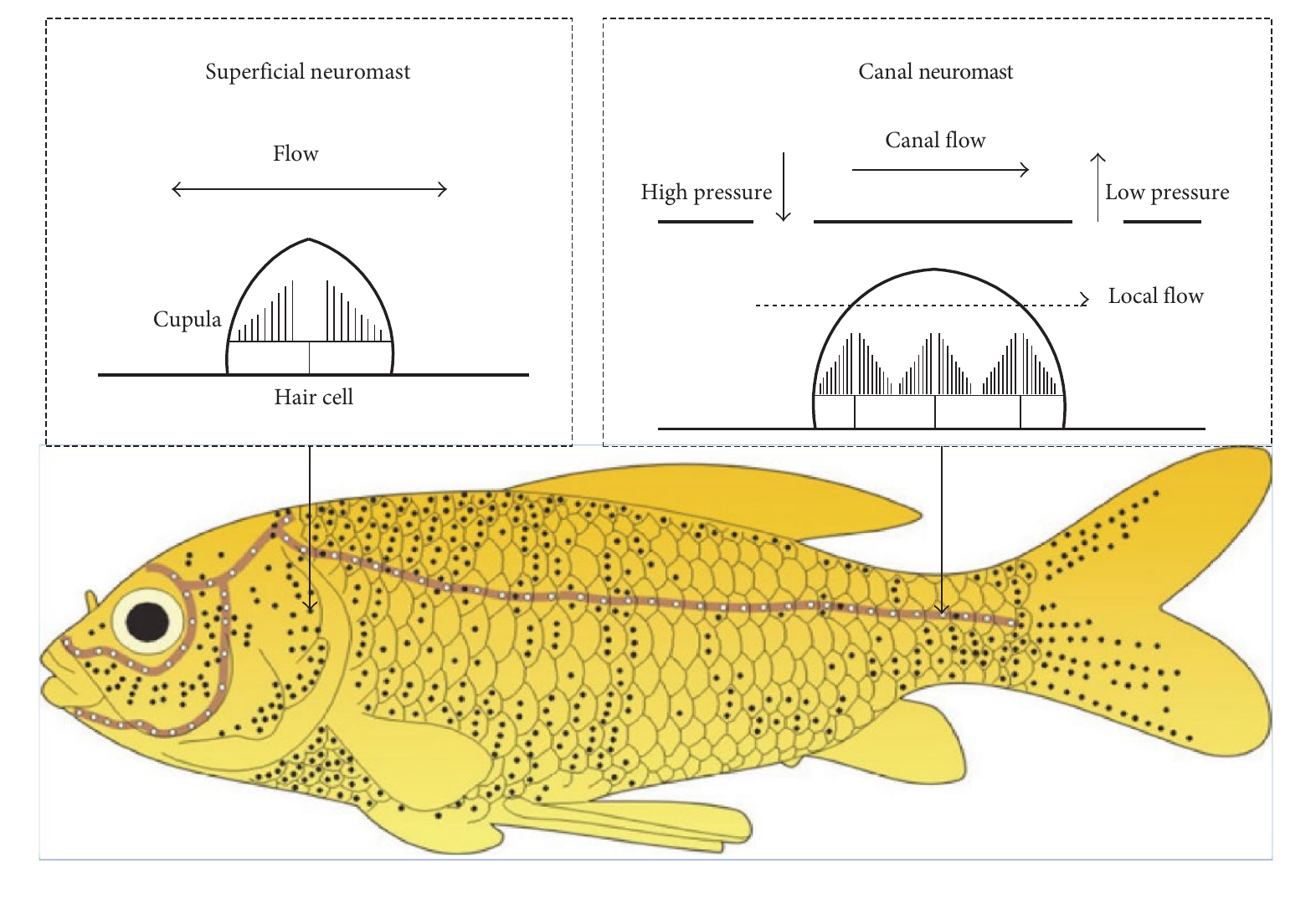}
	\caption{Lateral line and neuromasts of a fish. Black dots represent locations of SNs, and white dots show the approximate locations of canal pores. \cite{66}}
	\label{Fig 1}
\end{figure}

Lateral line is the most highly differentiated structure in the sensory organs of the skin, which is typically sulcus or tubular. It is a sensory organ peculiar to fish and aquatic amphibians and important for them to maneuver in the darkness \cite{63}. The sensory unit of the lateral line is the neuromast, a receptor that consists of sensory hair cells and support cells \cite{68}. There are two types of lateral line neuromasts: superficial neuromast (SN) and canal neuromast (CN), as shown in Figure \ref{Fig 1}. Both of these two kinds of neuromasts sense the stimulation generated by water flow through sensory cells. Due to the difference in distribution, number and morphology of sensory cells, the two kinds of neuromasts have different functions \cite{103}.

The SNs which act as displacement sensors are free-standing on the skin or on pedestals grown above the skin. They are usually located in lines on the fish body \cite{69}. The ability of sensing the flow direction and velocity is mainly realized by the SNs which are sensitive to the displacement and respond to the low-frequency direct current component. When the water flow and the fish surface move relatively to each other, the SNs bend, causing the neuromasts below to produce nerve impulses which will be transmitted from nerve endings to the nerve centers of the brain. Under such a mechanism, fish can sense the flow information with the help of SNs. On the other hand, the CNs are equivalent to pressure gradient sensors which can sense pressure gradient and are sensitive to acceleration and respond to high-frequency components. The CNs are located in the lateral line canals that are full of mucus under the epidermis of fish and communicate with the external water environment through some small holes \cite{104}. When there is a velocity gradient between adjacent holes, the pressure difference will be generated, leading to the fluid movement in the lateral line canals, which triggers the nerve impulse. The lateral line system composed of CNs and SNs can sense various stimuli from different directions, so that fish can acquire enough information for a full sense of surrounding water environment.

\begin{figure}[htbp]
	\centering
	\includegraphics[width=\columnwidth]{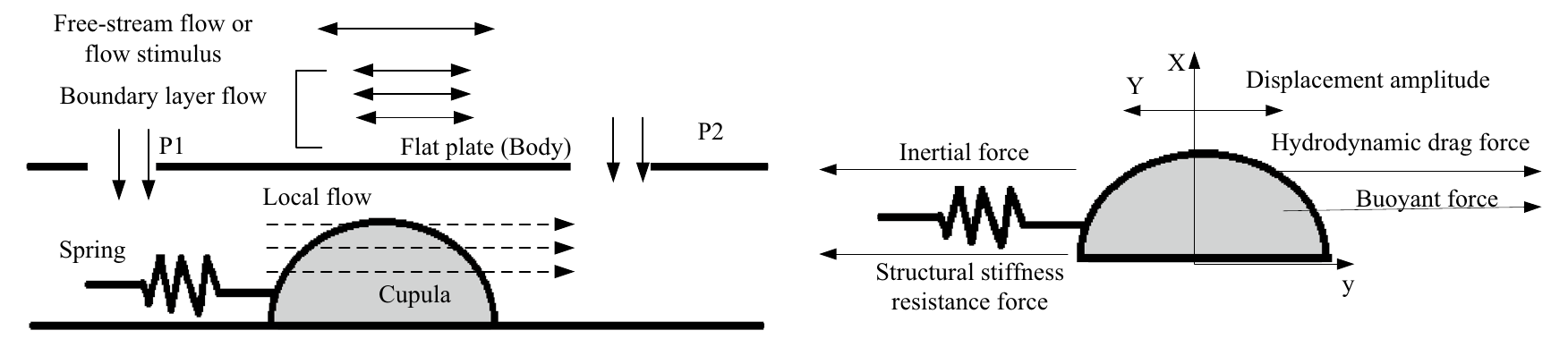}
	\caption{Biomechanical model and force analysis for a CN. \cite{67}}
	\label{Fig 2}
\end{figure}

Scientists have established theoretical models to describe the sensing mechanisms of these neuromasts and to study the interactions between flow and fish, which is instructive to the development of ALL system. As shown in Figure \ref{Fig 2}, the CN can be regarded as a rigid hemisphere sliding over a frictionless plate in biomechanics which is coupled with a linear spring \cite{67}\cite{71}. For the forces on the CNs, the governing equation can be described as:
\[F_m+F_k=F_u+F_b\]
The terms on the left are the inertial force and the structural stiffness resistance which can be written as the following forms respectively:
\[F_m=M\frac{d^2Y(t)}{dt^2}\]
\[F_k=KY(t)\]
where $M$ represents the cupula mass, $Y(t)$ represents the displacement induced by the passing flow, $K$ is the sliding stiffness. The terms on the right are the hydrodynamic drag force and buoyant force due to the pressure difference which can be defined as the following forms:
\[F_u=D[\frac{dY(t)}{dt}-\frac{dW(t)}{dt}]\]
\[F_b=M\frac{d^2W(t)}{dt^2}\]
where $D$ is the drag coefficient, $W(t)$ is the displacement of external flow. Thus $\frac{dY(t)}{dt}-\frac{dW(t)}{dt}$ represents the relative velocity of the cupula and external flow. According to the above results, the governing equation can be written as
\[M\frac{d^2Y(t)}{dt^2}+KY(t)=D[\frac{dY(t)}{dt}-\frac{dW(t)}{dt}]+M\frac{d^2W(t)}{dt^2}\]
In order to describe the sensitivity of CNs, considering that the equation is linear, the displacement can be decomposed into different frequencies. At a certain frequency, $Y(t)$ and $W(t)$ are the steady-state oscillatory displacement and the oscillating flow displacement with a form of $Y(t)=Y_0(f)e^{i2 \pi ft}$ and $W(t)=W_0(f)e^{i2 \pi ft}$. So the velocity of the flow is expressed as
\[V(t)=\frac{dW(t)}{dt}=i2 \pi fW_0(f)e^{i2 \pi ft}=V_0(f)e^{i2 \pi ft}\]
The frequency-dependent sensitivity of the CNs is defined as the ratio of displace amplitude of cupula and the velocity amplitude of flow, which is shown in the following formula:
\[S_{CN}(f)=\frac{Y_0(f)}{V_0(f)}=\frac{1}{-2\pi f_t}\frac{1+\frac{\sqrt{2}}{2}(1+i)\sqrt{\frac{f}{f_t}}+\frac{1}{3}i\frac{f}{f_t}}{N_r+i\frac{f}{f_t}-\frac{\sqrt{2}}{2}(1-i)({\frac{f}{f_t}})^{\frac{3}{2}}-\frac{1}{2}({\frac{f}{f_t}})^2}\]
where $f_t$ is the transition frequency expressed as $f_t=\frac{\mu}{2\pi \rho_wa^2}$ which determine viscous $(f<f_t)$ or inertial $(f>f_t)$ forces dominate the fluid forces applied on the cupula. $\mu$ is the dynamic viscosity of fluid. $N_r=\frac{Ka\rho_w}{6\pi \mu^2}$ which represents the resonance properties is the resonance factor.

\begin{figure}[htbp]
	\centering
	\includegraphics[width=8cm]{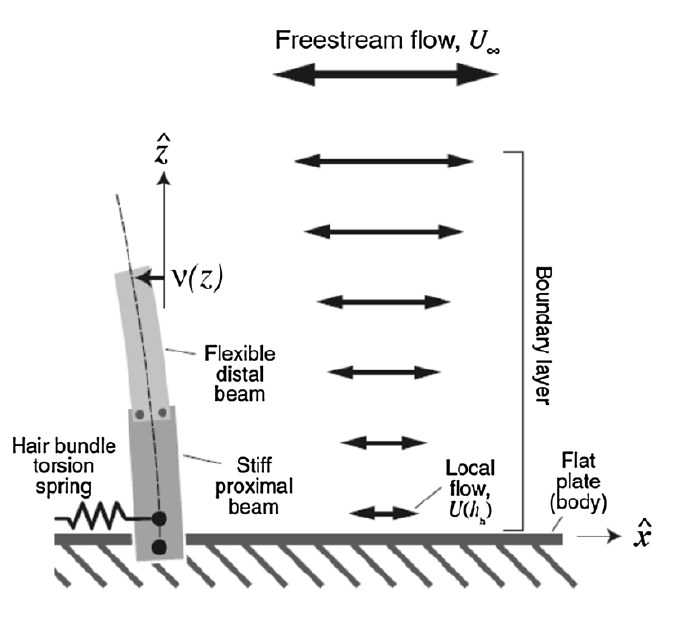}
	\caption{Biomechanical model of a SN. \cite{70}}
	\label{Fig 3}
\end{figure}

Different from the CN, the SN is modeled as two connecting beams with different bending rigidity, as shown in Figure \ref{Fig 3}. The distal beam is more flexible than the proximal beam because of the difference in material properties of the proximal and distal parts of cupula. A spring is used to simulate the torsional stiffness generated by the hair beam. The governing equation is different from that of CNs, forces applied on the beams are also functions of elevation $z$ due to the uneven distribution of forces when a beam bends. The equation is expressed as
\[F_m(z)+F_e(z)=F_u(z)+F_a(z)+F_b(z)\]
The terms from left to right represent the inertial force, elastic stiffness term, hydrodynamic drag force, acceleration reaction force and buoyant force respectively. Similar to CNs, the sensitivity of SNs is defined as the ratio of cupula deflection $v(H)$ at the height of the beam and free-stream velocity $U_\infty$, which is written as
\[S_{SN}(f)=\frac{v(H)}{U_\infty}=-\frac{ib_w}{2\pi fb_m}[1-\frac{i\pi fb_m\delta^4}{2EI+i\pi fb_m\delta^4}e^{-\frac{H(1+i)}{\delta}}]+\sum_{j=0}^{3}C_je^{i^jH^4\sqrt{\frac{2\pi fib_m}{EI}}}\]
where $C_j$ represents a sequence of four integration constants. $H$ is the height of the top of the beam. $EI$ is the bending modulus of the beam. $\delta$ is the thickness of boundary layer expressed as $\delta=\sqrt{\frac{2\mu}{\rho_w\omega}}$, where $\omega$ is the angular speed of the stimulus. More details on the parameters $b_m$ and $b_w$ can be referred to the paper \cite{70}.

\begin{figure}[htbp]
	\centering
	\includegraphics[width=\columnwidth]{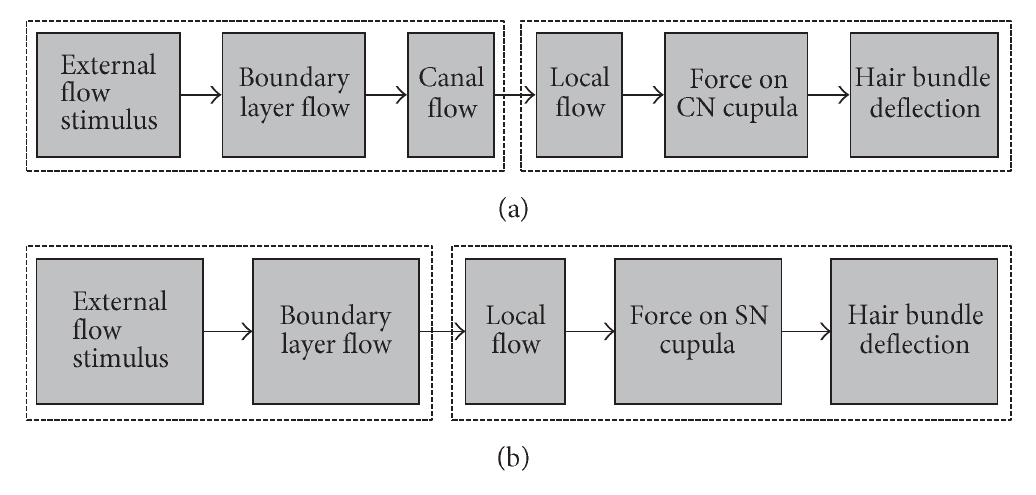}
	\caption{Propagation path of lateral line neuromasts. (a)Propagation path of CNs. (b)Propagation path of SNs. \cite{67}}
	\label{Fig 4}
\end{figure}

Figure \ref{Fig 4} shows the propagation path of lateral line neuromasts. On one hand, for CNs, the propagation path is divided into two steps. At the first step, the velocity or acceleration of the external free flow is converted into the velocity of the local flow with the help of the boundary layer and canals. Flow outside induces the pressure difference between the canals, which triggers flow velocity inside the canal. At the second step, canal flow applies fluid forces on the cupula and results in the CN deflection. On the other hand, for SNs, at the first step, the velocity or acceleration of the external free flow is converted into the velocity of the local flow similarly without the reflection of canal flow. At the second step, SNs deflects under the forces induced by local flow \cite{67}.

With the assistance of SNs and CNs, fish can detect the flow direction and speed and pressure gradients respectively. Moreover, SNs can distinguish fields in a spatial uniform flow and in a turbulent flow, while CNs only respond to a non-uniform flow field, such as the fluctuation of water produced by a vibrating sphere or a swimming fish.

\section{The existing ALL sensors and systems}

\subsection{ALL sensor unit}
As mentioned above, scientists have established mathematical models to interpret the mechanisms that how fish acquire fluid information assisted by lateral line. The results can be an inspiration of ALL. Owing to the limitations of existing technologies for underwater detection such as scattering and multipath propagation issues for acoustic sensors and turbidity of the sea for optical sensors \cite{67}, varieties of fish sensing organs inspired sensors were developed using different principles. Considering that commercial pressure sensors are not as sensitive as the lateral line receptors and the functions are limited, various kinds of self-developed ALL sensors have been explored. The existing ALL sensors are based on different sensing mechanisms, including piezoresistive, piezoelectric, capacitive, optical, thermal and magnetic effect. The research status of ALL sensors with different sensing mechanisms will be discussed below.

\subsubsection{Piezoresistive ALL Sensors}
\begin{figure}[htbp]
	\centering
	\includegraphics[width=14cm]{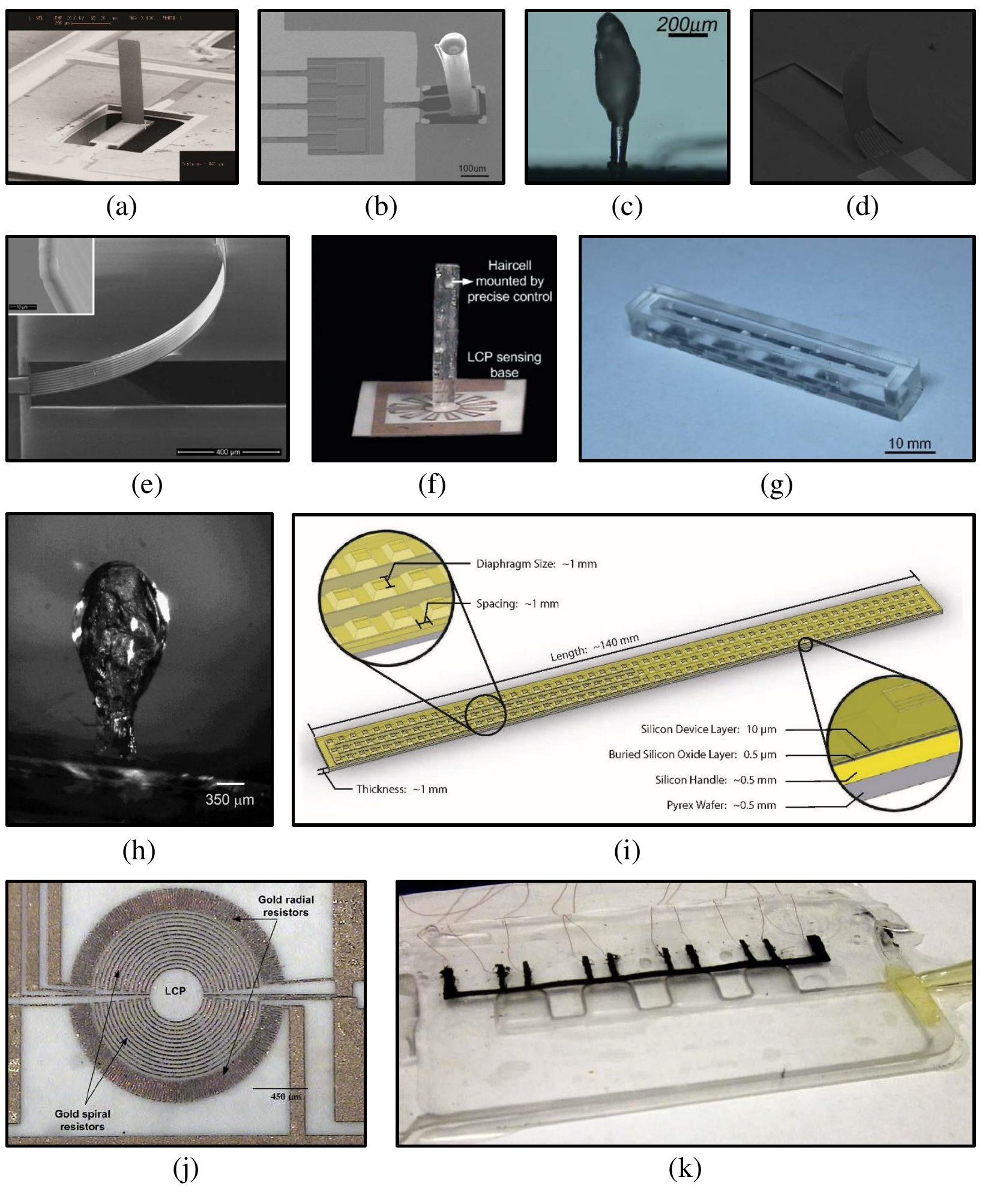}
	\caption{Various piezoresistive ALL sensors. (a)Scanning electron micrograph of
		a single artificial hair cell sensor \cite{1}. (b)Scanning electron micrograph of an AHC sensor \cite{3}. (c)The front-view of a hair sensor after being coated with the hydrogel material \cite{7}. (d)Scanning electron micrograph on a bent cantilever \cite{8}. (e)Scanning electron micrograph on a fabricated cantilever \cite{9}. (f)An angle view of the complete sensor with hair cell \cite{10}. (g)Photographic image of the final ALL canal system prototype \cite{12}. (h)Cupula formed using nanofibrils scaffold has a prolate spheroid shape \cite{11}. (i)Diagram of the pressure sensor array with basic structure depicted \cite{13}. (j)Photograph of the pressure sensor array \cite{14}. (k)Optical microscopic image of a fabricated full-bridge LCP sensor with two radial and two spiral gold piezoresistors \cite{15}.}
	\label{Fig 5}
\end{figure}

Piezoresistive sensor is a device based on the piezoresistive effect of the semiconductor material on the substrate. Piezoresistive effect refers to a phenomenon that the electrical resistance of the material changes while it is subjected to force. As a result, the bridge on the substrate produces corresponding unbalanced output. In this way, the substrate can be directly used as an element to measure pressure, tension and etc. And then based on the quantity measured directly, the information about the environment is available. Figure \ref{Fig 5} shows various piezoresistive ALL sensors mentioned below.

In 2002, Fan \emph{et al.} firstly made a major breakthrough in piezoresistive ALL sensors fabrication using combined bulk micromachining methods and an efficient three-dimensional assembly method named plastic deformation magnetic assembly (PDMA) process. They leveraged PDMA to realize the vertical cilium, which was important in the hair cell. A single sensor was composed of an in-plane fixed-free cantilever mainly made of Boron ion diffused Si using etching technology, a vertical artificial cilium attached at the free end and a strain gauge located at the base of the horizontal cantilever which was used to sense the bending of the vertical cilium. Subjected to the impact of local flow, the vertical cilium bent and transferred the influence to the cantilever beam. The corresponding results were measured by the strain gauge. The sensors were used to detect laminar flows ranging from 0.1 to 1 m/s \cite{1}.

In 2003, Chen \emph{et al.} compared the above sensor with the hot-wire anemometer and improved the design by rigidly connecting the vertical hair to the substrate and placing the strain gauge at the root of the hair directly, which made an advance in the spatial resolution of the sensor \cite{2}. In order to further improve the sensitivity and resolution, in 2007, Yang \emph{et al.} proposed another highly sensitive piezoresistive flow sensor fabricated on a Silicon-on-insulator (SOI) wafer, the cilium of which was made of photodefinable SU-8 epoxy and adopted a symmetric cylindrical shape. The sensor was used to detect the steady-state laminar flow and oscillatory flow with a threshold down to 0.7 mm/s \cite{3}\cite{4}\cite{5}. In 2010, they assembled piezoresistive sensors on the surface of a cylindrical polyvinyl chloride (PVC) model and put forward an adaptive beamforming algorithm in order to locate the dipole source \cite{6}.

To match the threshold sensitivity of the integrated fish flow sensory system, McConney \emph{et al.} created a bio-inspired hydrogel-capped hair sensory system in 2008 using a precision drop-casting method. They added extremely compliant and high-aspect-ratio hydrogel cupula (polyethylene glycol) to the SU-8 hair sensors, as a result of which, the sensitivity of the sensors was enhanced by about two orders of magnitude (2.5 $\mu$m/s) \cite{7}.

Unlike the sensors mentioned above whose material was mainly silicon, Qualtieri \emph{et al.} reported on a kind of ALL sensor in 2011, the key component of which was the stress-driven Aluminium Nitride (AlN) cantilevers. The structures utilizing a multilayered cantilever AlN/Molybdenum (Mo) and a Nichrome 80/20 alloy as piezoresistor were realized by means of micromachining techniques combining optical lithography and etching process. The piezoresistor showed a sensitivity to directionality and low value pressure with a detection threshold of 0.025 bar \cite{8}. Besides, in 2012, they deposited a water resistant parylene conformal coating on the hair cell and developed a biomimetic waterproof Si/SiN multilayered cantilever using surface micromachining techniques. The sensor showed mechanical robustness in high-speed flow and had the capability of discriminating the flow direction at low frequencies \cite{9}.

In 2014, Kottapalli \emph{et al.} developed an artificial SN sensor array composed of a liquid crystal polymer (LCP) membrane, a gold strain gauge and a Si-60 cilium fabricated by stereolithography with a high-aspect ratio of 6.5. The sensors demonstrated a high sensitivity of 0.9 mV/(m/s) and 0.022 V/(m/s) while detecting air and water flows. And the threshold velocity limits were 0.1 m/s and 15 mm/s, respectively \cite{10}. In 2016, they created a canopy-like nanofiber pyramid around the Si-60 polymer cilium and then dropped the casting hydrogel cupula onto the nanofiber scaffold to enhance the sensors. The velocity detection threshold of the sensor was 18 mm/s while it is used in water flow sensing \cite{11}.

All sensors mentioned above feature a cilium and a cantilever beam, which bends under the impact of water flow and is sensitive to flow velocity. Except for this, other piezoresistive sensors are mostly planar and the piezoresistors are directly installed on the substrate to detect underwater pressure distribution and variations.

In 2017, Jiang \emph{et al.} integrated cantilevered flow-sensing elements mainly made of polypropylene and polyvinylidene fluoride (PVDF) layers in a polydimethylsiloxane (PDMS) canal and used it to detect a dipole vibration source. The sensors showed high-pass filtering capability and a pressure gradient detection limit of 11 Pa/m at the frequency of 115 Hz \cite{12}.

Additionally, Vicente \emph{et al.} used an array of off-the-shelf pressure sensors to detect cylindrical obstacles of round and square in 2007. The array consisted of hundreds of micro-electromechanical systems (MEMS) pressure sensors which were fabricated on etched Silicon and Pyrex wafers. A strain gauge was mounted on a flexible diaphragm which was a thin (20 $\mu$m) layer of Silicon attached at the edges of a square Silicon cavity with a width of 2000 $\mu$m and served as the sensing element with a pressure detection threshold of 1 Pa \cite{13}. In 2012, they presented a 1-D array of four sensors with a 15-mm center-to-center spacing. Each sensor had two key components: a strain-concentrating PDMS diaphragm and a resistive strain gauge made of a conductive carbon-black PDMS composite. And the resolution of it was 1.5 Pa \cite{14}.

To perform underwater surveillance, Kottapalli \emph{et al.} developed an array of polymer MEMS pressure sensors fabricated with a Cr (20 nm)/Au (700 nm) thick gold layer sputtered on a flexible substrate and LCP serving as the sensing membrane material. Installed on curved surfaces of the underwater vehicle bodies, the sensor detected underwater objects by sensing the pressure variations. Compared with Silicon-based hair vertical structures or thin metal cantilever beams, it showed a better sensitivity of 14.3 $\mu$V/Pa and a better resolution of 25 mm/s in water flow sensing \cite{15}.

\subsubsection{Piezoelectric ALL Sensors}
\begin{figure}[htbp]
	\centering
	\includegraphics[width=\columnwidth]{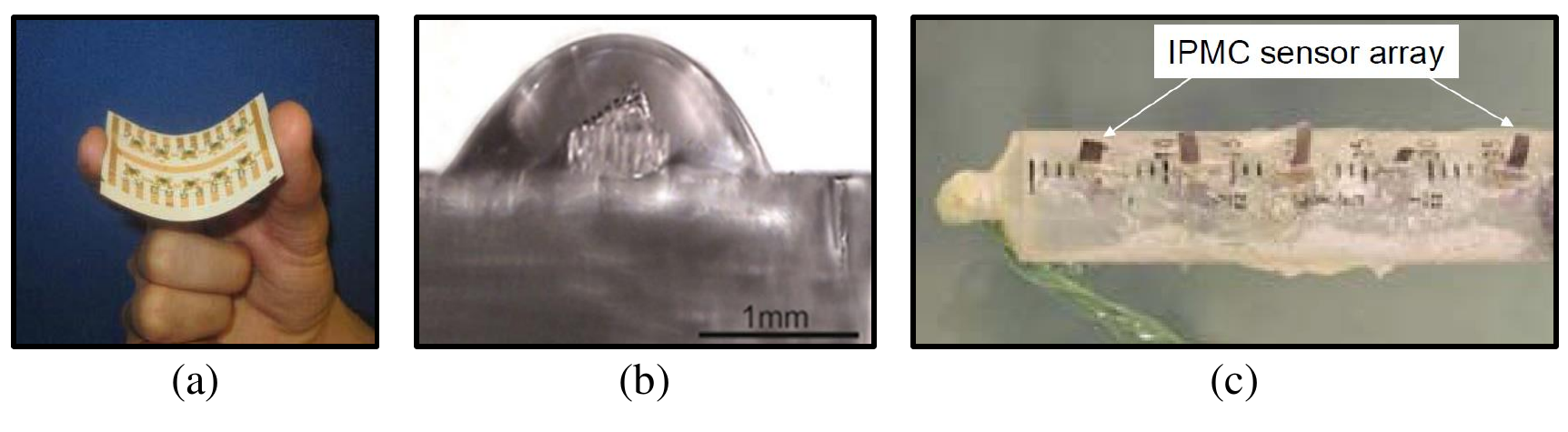}
	\caption{Various piezoelectric ALL sensors. (a)Array of 2 by 5 piezoelectric sensors on flexible LCP substrate \cite{16}. (b)Microscopic side-view image of the sensor showing the hydrogel cupula and the PDMS pillars with height gradient \cite{18}. (c)The IPMC-based lateral line prototype \cite{19}.}
	\label{Fig 6}
\end{figure}

Piezoelectricity refers to the electric charge generated on the surface of some certain materials while subjected to external forces. This effect inspires another kind of ALL sensors which is able to sense environment by collecting the electric information. Figure \ref{Fig 6} shows various piezoelectric ALL sensors mentioned below.

For the performance of situational awareness and obstacle avoidance, Asadnia \emph{et al.} used floating bottom electrode to design an array of Pb(Zr$_{0.52}$Ti$_{0.48}$)O$_3$ thin-film piezoelectric pressure sensors in 2013 \cite{16}. Packaged into an array of 25 sensors on a flexible liquid crystal polymer substrate patterned with gold interconnects, the array was used to locate a vibrating sphere dipole in water and showed a resolution of 3 mm/s in detecting oscillatory flow velocity. Besides, the sensors had many advantages such as self-powered, miniaturized, light-weight, low-cost and robust. In 2015, they optimized the sensor by mounting a stereolithographically fabricated polymer hair cell on microdiaphragm with floating bottom electrode. The sensors demonstrated a high-pass filtering nature with a cut-off frequency of 10 Hz, a high sensitivity of 22 mV/(mm/s) and a resolution of 8.2 mm/s in water flow detection \cite{17}. In 2016, they reported the development of a new class of miniature all-polymer flow sensors with an artificial ciliary bundle fabricated by combining bundled PDMS micro-pillars with graded heights and electrospinning PVDF piezoelectric nanofiber tip links. By means of precision drop-casting and swelling processes, a dome-shaped hyaluronic acid hydrogel cupula encapsulating the artificial hair cell bundle was formed. The sensors achieved a sensitivity of 300 mV/(m/s) and a threshold detection limit of 8 $\mu$m/s respectively \cite{18}.

In 2011, Abdulsadda \emph{et al.} proposed a novel ALL sensors utilizing the inherent sensing capability of ionic polymer-metal composites (IPMCs). An IPMC consisted of three layers, with an ion-exchange polymer membrane sandwiched by metal electrodes. Detectable electrical signals were produced under the impact of external forces. The IPMC flow sensor were used to localize dipole sources 4-5 body away and demonstrated a threshold detection limit of less than 1 mm/s \cite{19}.

\subsubsection{Capacitive ALL Sensors}
\begin{figure}[htbp]
	\centering
	\includegraphics[width=8cm]{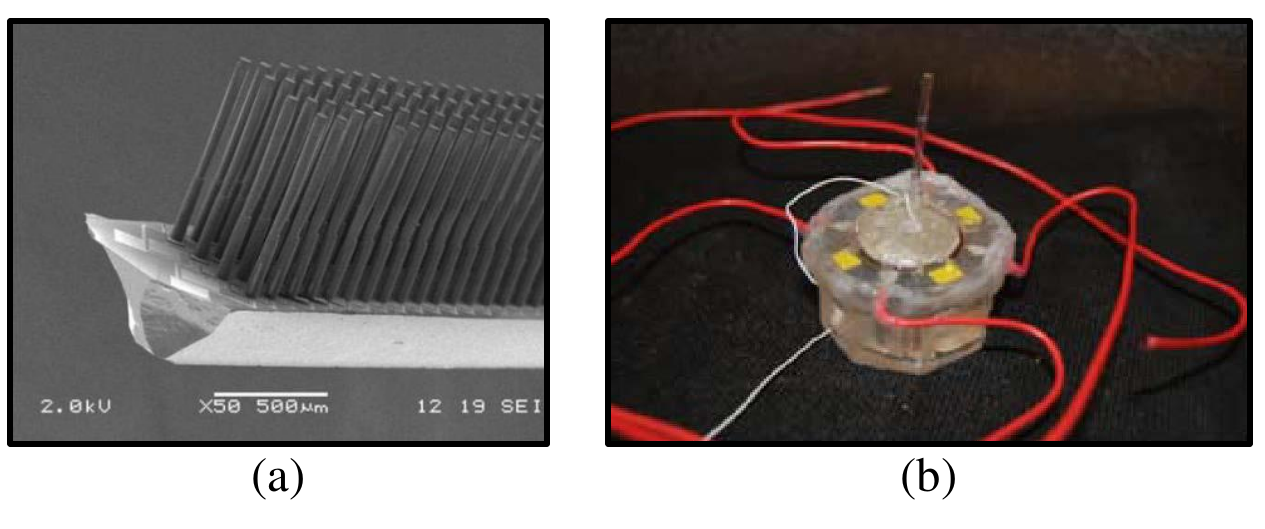}
	\caption{Various capacitive ALL sensors. (a)Scanning Electron Microscope image of actual sensors \cite{20}. (b)Top views of the ALL sensor \cite{23}.}
	\label{Fig 7}
\end{figure}

Owing to the high sensitivity and low power consumption, capacitance principle has been widely used in many different types of sensors. The key component is the capacitive readout which has the capability of converting external stimulus into capacitance changes, which provides an effective way to detect underwater pressure and flow velocity. Like the piezoresistive sensors, the hair attached to the membrane will respond to the impact from the local flow. And then the membrane reflects and changes the distance or gap between the electrodes. As a result, the change in capacitance is quantitatively related to the external impact. Figure \ref{Fig 7} shows various capacitive ALL sensors mentioned below.

In 2007, Krijnen \emph{et al.} reported developments in hair sensors based on mechanoreceptive sensory hairs of crickets using artificial polysilicon technique to form Silicon Nitride-suspended membranes and SU-8 polymer processing for hairs with diameters of about 50 $\mu$m and up to 1 mm in length. The membranes havd thin chromium electrodes on top which formed variable capacitors and the sensitivity of the sensor is 1.39 pF/rad \cite{20}\cite{22}. In 2010, they realized the dense arrays of fully supported flexible SU-8 membranes with integrated electrodes underneath that supported cylindrical hair-like structures on the top. While used in air flow detection, the mechanical sensitivity at the frequency of 115 Hz was 0.004 rad/(m·s) \cite{21}.

Another capacitive whisker sensor inspired by seal vibrissae was developed by Stocking \emph{et al.} to measure the flow velocity and detect the direction in 2010. They mounted a rigid artificial whisker on a novel cone-in-cone parallel-plate capacitor base which was covered by a PDMS membrane. Numerical simulation predicted the change of capacitor output signal in a range of 1 pF when the flow velocity varied from 0 to 1.0 m/s \cite{23}.

\subsubsection{Optical ALL Sensors}
\begin{figure}[htbp]
	\centering
	\includegraphics[width=\columnwidth]{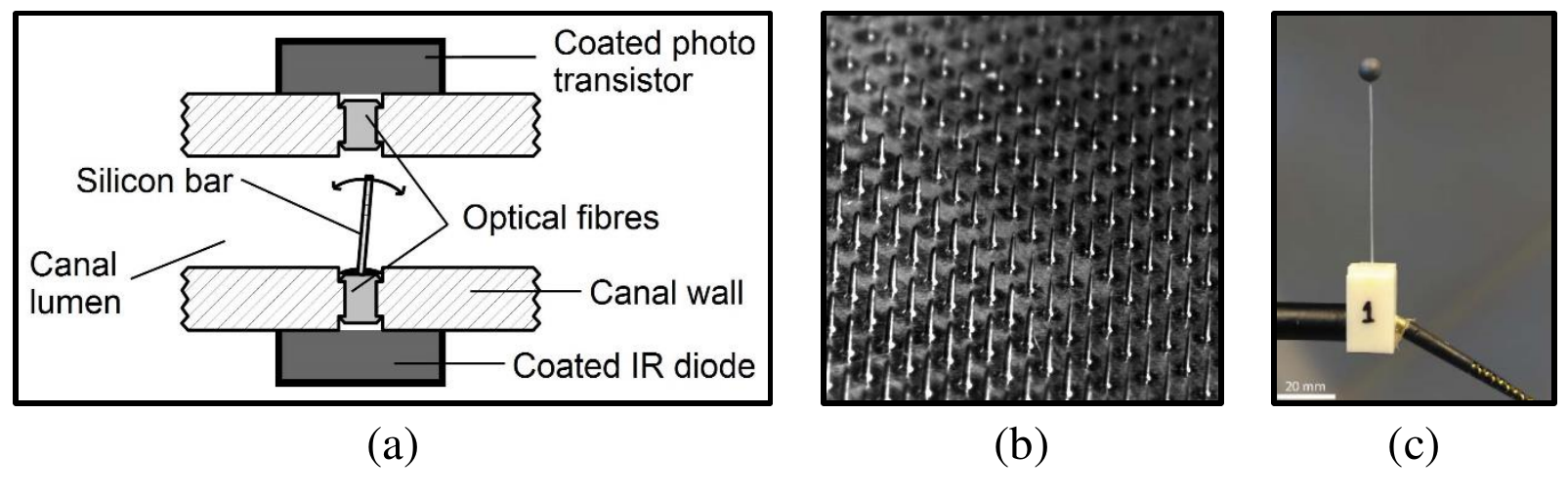}
	\caption{Various optical ALL sensors. (a)Scheme of an artificial CN \cite{24}. (b)Scanning-Electron Microscope image of a pillar array \cite{25}. (c)The photograph of the all-optical sensor \cite{26}.}
	\label{Fig 8}
\end{figure}

Optical principles have also been used to develop ALL sensors. Figure \ref{Fig 8} shows various optical ALL sensors mentioned below. Klein \emph{et al.} made a great breakthrough in this area in 2011. The artificial canal neuromasts segment they developed consisted of a transparent silicone bar which had the same density of water and an infrared light emitting diode at one end of the silicone bar. To detect the fluid motion, light, leaving the opposite end of the silicone bar, illuminated an optical fiber that was connected to a surface mounted devices(SMD) phototransistor and the output was amplified and converted to be stored on a computer. The sensors are used to detect water movements caused by a stationary vibrating sphere or a passing object and vortices caused by an upstream cylinder. According to the acquired information, they calculated the bulk flow velocity and the size of the cylinder producing the vortices. The detection limit in water flow was 100 $\mu$m/s \cite{24}.

Another method was put forward by Wolfgang \emph{et al.} in 2009. The key component of the sensor was flexible micro-pillars which protruded into local flows and bent subjected to exerted drag forces. The pillar was fabricated from the elastomer PDMS and the deflection was measured by means of optical methods \cite{25}.

In 2018, Wolf \emph{et al.} presented an all-optical 2D flow velocity sensor consisting of optical fibres inscribed with Bragg gratings supporting a fluid force recipient sphere. The artificial neuromast demonstrated a threshold of 5 mm/s at a low frequency and 5 $\mu$m/s at resonance with a typical linear dynamic range of 38 dB at 100 Hz sampling. Additionally, the artificial neuromast is capable of detecting flow direction within a few degrees \cite{26}.

\subsubsection{Hot-Wire ALL Sensors}
\begin{figure}[htbp]
	\centering
	\includegraphics[width=8cm]{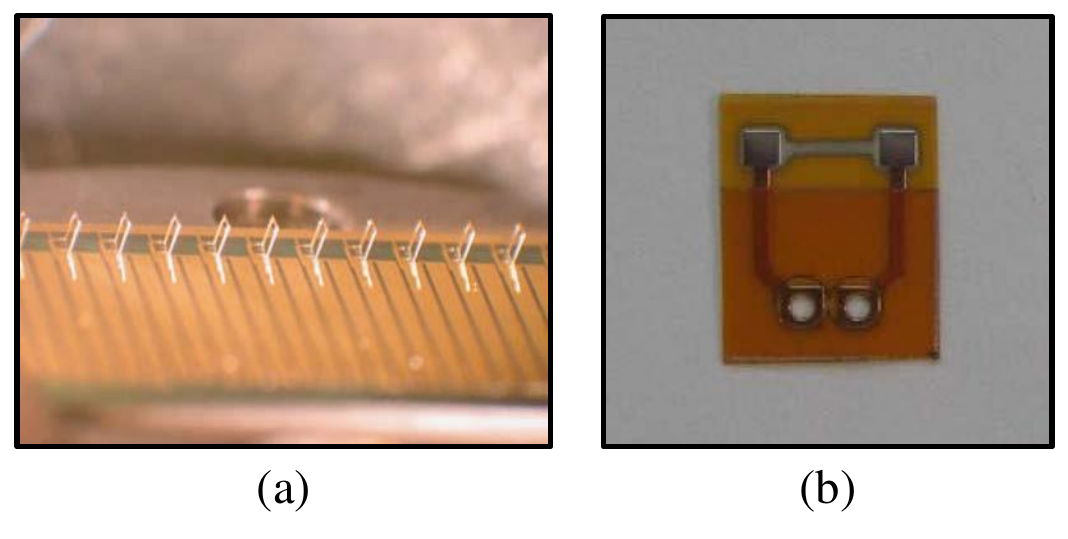}
	\caption{Various hot-wire ALL sensors. (a)An optical micrograph of an ALL \cite{28}. (b)A sensor unit \cite{30}.}
	\label{Fig 9}
\end{figure}

Hot-wire anemometer (HWA) uses a heated wire placed in the air. While air or water flows through it, heat loss leads to changes in temperature and resistance. Therefore, we can measure the velocity by detecting electrical signals. Figure \ref{Fig 9} shows various hot-wire ALL sensors mentioned below.

In 2006, Yang \emph{et al.} developed a surface-micromachined, out-of-plane ALL sensor array using the principle of thermal HWA. Inspired by the SNs of fish, the hot wire was lifted from the substrate by two pointed heads. They utilized photolithography technique to fabricate the sensor in plane and assembled it outside by three-dimensional magnetic assembly. The sensor afterwards was used to track the position of a vibrating dipole source and exhibited a threshold of 0.2 mm/s with a bandwidth of 1 kHz \cite{27}\cite{28}\cite{29}.

Liu \emph{et al.} proposed a novel micromachined hot-film flow sensor system realized by using a film depositing process and a standard printed circuit in 2009. They preprinted the sensor electrodes and electronic circuits on a flexible substrate of polyimide and utilized Cr/Ni/Pt as the sensing element with a resistance temperature coefficient around 2000 ppm/K. The resolution of this sensor was 0.1 m/s \cite{30}.

\subsection{ALL sensors placement optimization}
With the development of ALL sensors unit, more efforts have been devoted to the sensors placement optimization in order to simulate the fish lateral line to a greater extent. Verma \emph{et al.} used a larva-shaped swimmer exposed in disturbances induced by oscillating, rotating and cylinders to conduct experiments in 2020. Combining Navier-Stokes equations with Bayesian experimental design and with a purpose of detecting the location of the source, they presented that shear sensors should be installed on the head and the tail while pressure sensors should be distributed uniformly along the body and intensively on the head, which is similar to real fish lateral line \cite{105}. In 2019, Xu \emph{et al.} put forward an optimal weight analysis algorithm combined with feature distance and variance evaluation and 3 indexes to evaluate the performance of the sensor array. They also briefly discussed the optimal number of sensors \cite{106}. This work has provided new ideas for studies in ALL sensors distribution optimization in the future.

~\\
In this section, we have presented ALL sensors based on different sensing mechanisms and briefly introduced works in sensors placement optimization. Table 2 summarizes the parameters of the ALL sensors, including transduction mechanism, processing technique, material and sensitivity. Though great progress has been made in the design and fabrication of ALL sensors, the sensors mentioned above are mostly simple imitations of the real fish lateral line and have a long way to go in sensitivity. Firstly, the sensitivity of a single sensory unit can be further improved. Additionally, the neuromasts distribution of fish is continuously optimized in evolution. Thus, it is necessary for us to find out the optimal distribution of ALL sensors according to different shapes of robotic fish in order to simulate the lateral line to the maximum extent possible and develop a complete ALL system based on existing local sensor arrays. Last but not least, we need more efficient signal processing methods to take full advantage of the information acquired by the ALL system and make decisions like a living body. ALL system like this has great potential for future oceanographic research.

\begin{longtable}[]{|c|c|c|c|c|}
	\caption{Summary of ALL sensors}\\
	\hline
	\begin{tabular}[c]{@{}c@{}}Transduction\\ mechanism\end{tabular} & Author                                                                                                                                                                                                                            & \begin{tabular}[c]{@{}c@{}}Processing\\ technique\end{tabular}                                               & Material                                                                                                                         & Sensitivity                                                                                                            \\ \hline
	\multirow{10}{*}{Piezoresistive}                                 & \begin{tabular}[c]{@{}c@{}}Fan et al.\\ 2002 \cite{1}\\ Chen et al\\ 2003 \cite{2}\end{tabular}                                                                                                 & \begin{tabular}[c]{@{}c@{}}PDMA for\\ vertical cilium,\\ micromachining\end{tabular}                         & \begin{tabular}[c]{@{}c@{}}Boron ion diffused\\ Si for piezoresistor,\\ Metal-permalloy\\ for hair\end{tabular}              & 100 mm/s                                                                                                               \\ \cline{2-5}
	& \begin{tabular}[c]{@{}c@{}}Chen et al.\\ 2007 \cite{3}\\ Yang et al.\\ 2007 \cite{4}\\ Chen et al.\\ 2006 \cite{5}\\ Yang et al.\\ 2010 \cite{6}\end{tabular} & \begin{tabular}[c]{@{}c@{}}Ion implantation,\\ deep reactive \\ ion etching\end{tabular}                     & \begin{tabular}[c]{@{}c@{}}Boron ion diffused\\ Si for piezoresistor,\\ SU-8 epoxy\\ for hair\end{tabular}                    & 0.1 mm/s                                                                                                               \\ \cline{2-5}
	& \begin{tabular}[c]{@{}c@{}}McConney et\\ al. 2008 \cite{7}\end{tabular}                                                                                                                                          & \begin{tabular}[c]{@{}c@{}}Photo\\ polymerization\end{tabular}                                               & \begin{tabular}[c]{@{}c@{}}Boron ion diffused\\ Si for piezoresistor,\\ SU-8-hydrogel\\ for hair\end{tabular}                 & 2.5 $\mu$m/s                                                                                                         \\ \cline{2-5}
	& \begin{tabular}[c]{@{}c@{}}Qualtieri et\\ al. 2011 \cite{8}\end{tabular}                                                                                                                                         & Micromachining                                                                                               & \begin{tabular}[c]{@{}c@{}}Aluminum Ni for\\ piezoresistor,\\ Nichrome alloy for hair\end{tabular}                            & 0.025 bar                                                                                                              \\ \cline{2-5}
	& \begin{tabular}[c]{@{}c@{}}Qualtieri et\\ al. 2012 \cite{9}\end{tabular}                                                                                                                                         & Micromachining                                                                                               & \begin{tabular}[c]{@{}c@{}}Si/SiN for\\ piezoresistor,\\ Parylene for hair\end{tabular}                                       & 50 mm/s                                                                                                                \\ \cline{2-5}
	& \begin{tabular}[c]{@{}c@{}}Kottapalli et\\ al. 2014 \cite{10}\\ Kottapalli et\\ al. 2016 \cite{11}\end{tabular}                                                                                 & \begin{tabular}[c]{@{}c@{}}Deep reactive \\ ion etching,\\ electrostatic\\ spinning\end{tabular}             & \begin{tabular}[c]{@{}c@{}}LCP for membrane,\\ gold for strain gauge,\\ Si-60 for hair,\\ HA-MA hydrogel for cupula\end{tabular} & \begin{tabular}[c]{@{}c@{}}100 mm/s (air),\\ 18 mm/s (water\\ flows)\end{tabular}                                        \\ \cline{2-5}
	& \begin{tabular}[c]{@{}c@{}}Jiang et al.\\ 2017 \cite{12}\end{tabular}                                                                                                                                            & Micromachining                                                                                               & \begin{tabular}[c]{@{}c@{}}PDMS for canal,\\ Polypropylene and\\ PVDF for piezoresistor\end{tabular}                            & 11 Pa/m                                                                                                                \\ \cline{2-5}
	& \begin{tabular}[c]{@{}c@{}}Vicente et\\ al. 2007 \cite{13}\end{tabular}                                                                                                                                          & Micromachining                                                                                               & Si                                                                                                                               & 1 Pa/m                                                                                                                 \\ \cline{2-5}
	& \begin{tabular}[c]{@{}c@{}}Vicente et\\ al. 2012 \cite{14}\end{tabular}                                                                                                                                          & Micromachining                                                                                               & \begin{tabular}[c]{@{}c@{}}PDMS for diaphragm,\\ a conductive\\ carbon-black PDMS\\ composite for strain gauge\end{tabular}      & 1.5 Pa                                                                                                                 \\ \cline{2-5}
	& \begin{tabular}[c]{@{}c@{}}Kottapalli et\\ al. 2012 \cite{15}\end{tabular}                                                                                                                                       & Micromachining                                                                                               & \begin{tabular}[c]{@{}c@{}}LCP for membrane,\\ gold for piezoresistors\end{tabular}                                              & 25 mm/s                                                                                                                \\ \hline
	\multirow{3}{*}{Piezoelectric}                                   & \begin{tabular}[c]{@{}c@{}}Asadnia et\\ al. 2013 \cite{16}\\ Asadnia et\\ al. 2015 \cite{17}\end{tabular}                                                                                       & \begin{tabular}[c]{@{}c@{}}Micromachining\\ and the sol-gel\\ method\end{tabular}                            & \begin{tabular}[c]{@{}c@{}}Pb(Zr$_{0.52}$Ti$_{0.48}$)O$_3$\\ for membrane,\\ Si-60 for hair\end{tabular}                        & 3 mm/s                                                                                                                 \\ \cline{2-5}
	& \begin{tabular}[c]{@{}c@{}}Asadnia et\\ al. 2016 \cite{18}\end{tabular}                                                                                                                                          & \begin{tabular}[c]{@{}c@{}}Precision\\ drop-casting\\ and swelling\\ processes\end{tabular}                  & \begin{tabular}[c]{@{}c@{}}PDMS for micro-pillars,\\ PVDF for tip links,\\ HA-MA hydrogel for cupula\end{tabular}                & 8 $\mu$m/s                                                                                                           \\ \cline{2-5}
	& \begin{tabular}[c]{@{}c@{}}Abdulsadda et\\ al. 2011 \cite{19}\end{tabular}                                                                                                                                       & Micromachining                                                                                               & IPMC                                                                                                                             & \begin{tabular}[c]{@{}c@{}}Localization\\ accuracy at the\\ source-sensor\\ separation of\\ 1 body length\end{tabular} \\ \hline
	\multirow{2}{*}{Capacitive}                                      & \begin{tabular}[c]{@{}c@{}}Krijnen et\\ al. 2007 \cite{20}\\ Krijnen et\\ al. 2010 \cite{21}\\ Krijnen et\\ al. 2003 \cite{22}\end{tabular}                                    & \begin{tabular}[c]{@{}c@{}}Sacrificial\\ poly-silicon\\ technology,\\ SU-8 polymer\\ processing\end{tabular} & \begin{tabular}[c]{@{}c@{}}Silicon-nitride\\ for membranes,\\ SU-8 polymer for hair\end{tabular}                                 & 0.004 rad/(m·s)                                                                                                 \\ \cline{2-5}
	& \begin{tabular}[c]{@{}c@{}}Stocking et\\ al. 2010 \cite{23}\end{tabular}                                                                                                                                         & Micromachining                                                                                               & PDMS for membrane                                                                                                                & N/A                                                                                                                    \\ \hline
	\multirow{3}{*}{Optical}                                         & \begin{tabular}[c]{@{}c@{}}Klein et al.\\ 2011 \cite{24}\end{tabular}                                                                                                                                            & N/A                                                                                                          & \begin{tabular}[c]{@{}c@{}}Silicone for transparent\\ bar, an infrared light\\ emitting diode\end{tabular}                       & 100 $\mu$m/s                                                                                                         \\ \cline{2-5}
	& \begin{tabular}[c]{@{}c@{}}Sebastian et\\ al. 2009 \cite{25}\end{tabular}                                                                                                                                        & N/A                                                                                                          & PDMS for the pillar                                                                                                              & N/A                                                                                                                    \\ \cline{2-5}
	& \begin{tabular}[c]{@{}c@{}}Wolf et al.\\ 2018 \cite{26}\end{tabular}                                                                                                                                             & N/A                                                                                                          & N/A                                                                                                                              & \begin{tabular}[c]{@{}c@{}}5 $\mu$m/s\\(resonance)\\and 5 mm/s\\(low frequency) \end{tabular}                                                                                                                 \\ \hline
	\multirow{2}{*}{Hot-Wire}                                        & \begin{tabular}[c]{@{}c@{}}Yang et al.\\ 2006 \cite{27}\\ Pandya et al.\\ 2006 \cite{28}\\ Chen et al.\\ 2006 \cite{29}\end{tabular}                                           & \begin{tabular}[c]{@{}c@{}}PDMA for\\ vertical cilium,\\ micromachining\end{tabular}                         & \begin{tabular}[c]{@{}c@{}}Pt/Ni/Pt film for\\ the thermal element,\\ polyimide for\\ support beams\end{tabular}                 & 0.2 mm/s                                                                                                                    \\ \cline{2-5}
	& \begin{tabular}[c]{@{}c@{}}Liu et al.\\ 2009 \cite{30}\end{tabular}                                                                                                                                              & \begin{tabular}[c]{@{}c@{}}film depositing\\process and\\standard printed\\ circuit\end{tabular}                                                                                                          & \begin{tabular}[c]{@{}c@{}}polyimide for\\substrate,\\Cr/Ni/Pt for\\sensing element\end{tabular}                                                                                                                              & 0.1 m/s                                                                                                                   \\ \hline
\end{longtable}

\section{Hydrodynamic environment sensing and vortices detection}
\begin{figure}[htbp]
	\centering
	\includegraphics[width=\columnwidth]{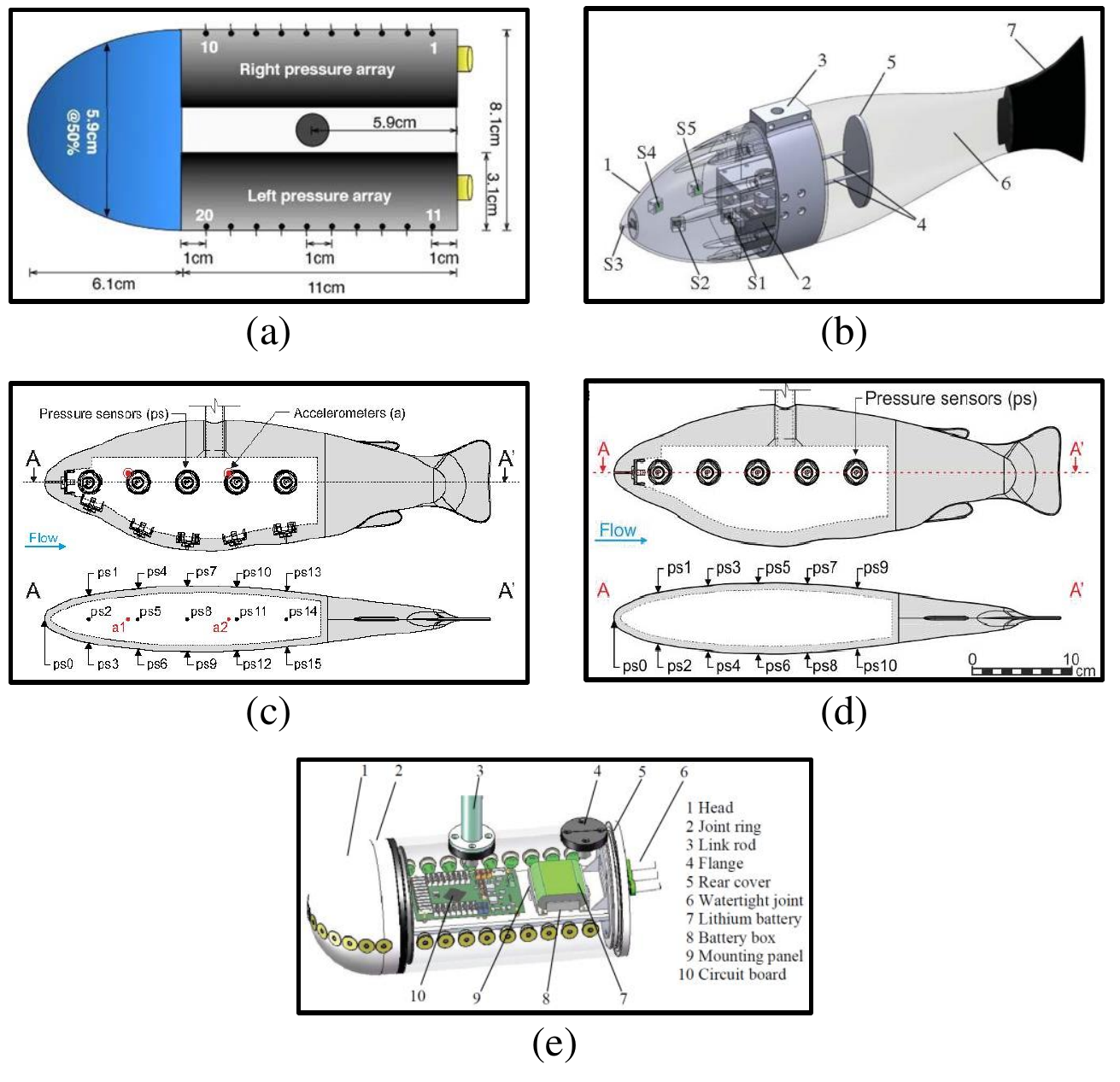}
	\caption{Different carriers boarded with ALL mentioned in Section 4. (a)A schematic diagram of the sensor platform used in \cite{49}. (b)CAD view of the robot used in \cite{73}: 1, rigid head of the robot; 2, servo-motor; 3, middle part for holding the head and the tail; 4, steel cables; 5, actuation plate; 6, compliant tail; 7, rigid fin; S1-S5, pressure sensors. (c)Location of the 16 pressure sensors and 2 accelerometers in the ALL Probe used in \cite{79}. (d)Illustration of the lateral line probe used for field measurements showing shape and sensor distribution in \cite{89}. (e)A 3-D model of the carrier in \cite{52}.}
	\label{Fig 10}
\end{figure}

Due to the limitations of existing measurements methods in complex natural flows and the emergence of the above mentioned various types of sensors, scientists have started to use ALL systems consisting of these sensors to obtain information from fluid environment. Efforts are devoted into the following fields: flow field characteristics identification, flow velocity and direction detection, vortex street properties detection. Carriers boarded with ALL mentioned in this section are shown in Figure \ref{Fig 10}.

\subsection{Flow field characteristics identification}
In 2013, Kruusmaa \emph{et al.} presented that the robotic fish boarded with 5 pressure sensors (Figure \ref{Fig 10}(b)) was able to identify the flow regimens (uniform flow and periodic turbulence) according to the pressure around the underwater vehicle in 2013 \cite{73}.

In 2018, they used the ALL probe consisting of 11 piezoresistive pressure sensors (Figure \ref{Fig 10}(d)) to capture hydrodynamic information. The probe provided an average flow velocity in turbulent flows which was comparable to the results measured practically. According to the characteristics of the fluid, the flows were classified by comparing the probability distribution of turbulent pressure fluctuation \cite{89}.

Liu \emph{et al.} also made great contribution in flow field characteristics identification. In 2019, they proposed an improved pressure distribution model to calculated the pressure around the ALL which consisting of 23 pressure sensors (Figure \ref{Fig 10}(e)) and then established a visualized pressure difference matrix to identify flow field in different conditions. A four-layer convolutional neural network model was constructed to evaluate the accuracy of this method \cite{52}.

\subsection{Flow velocity and direction detection}
In 2013, Kruusmaa \emph{et al.} proved that the flow speed can be estimated only by the average pressure on the sides of the robotic fish (Figure \ref{Fig 10}(b)). They put forward a fitted formula representing the relationship between the average pressure drop and the flow speed based on Bernoulli formula. Besides, they also reported that the flow direction could be detected because that the pressure on the side which the probe turned towards the flow was higher \cite{73}.

In the same group, they presented a new way for flow speed estimation with the help of ALL probe consisting of 10 piezoresistive pressure sensors (Figure \ref{Fig 10}(c)) without sensor calibration in 2015, which is of great convenience. Induced by the interactions between fluid and the robot body, fluctuations in the pressure field around the body could be detected by the ALL probe. Based on Bernoulli formula likewise, they introduced a semiempirical resampling process. Compared with results measured by an acoustic Doppler velocimeter in a vertical slot fishway, the accuracy of this method was validated \cite{79}.

Besides, Tuhtan \emph{et al.} made great progress in natural flow measurements using an ALL probe combined with signal processing methods in 2016. The probe is the same as showed in Figure \ref{Fig 10}(c). They proved that information acquired by the probe was transformed into two important hydrodynamic primitives, bulk flow velocity and bulk flow angle via canonical signal transformation and kernel ridge regression. Moreover, they showed that this method was effective not only when the sensor was parallel to the flow, but also in the condition that the angular deviation was large. While used in a natural river environment, the method had an error of 14 cm/s \cite{86}. In 2016, they presented a new method to estimate the flow velocity ranging from 0 to 1.5 m/s. They collected time-averaged flow velocity and pressure acquired by the ALL in highly turbulent flow and put forward a signal processing approach combining Pearson product-moment correlation coefficient features and artificial neural network \cite{47}. This method is potential to interpret the underwater preferences of fish in real environment.

Additionally, Liu \emph{et al.} in 2020, based on a fitting method and a back propagation neural network model, successfully predicted the flow velocity and direction and the moving velocity \cite{56}. This progress provided another possibility to identify hydrodynamic information for further study.

\subsection{Vortex street properties detection}
In 2006, Yang \emph{et al.} used ALL system to study the spatial velocity distribution of K{\'a}rm{\'a}n vortex street and visualized the velocity distribution of K{\'a}rm{\'a}n vortex street generated by a cylinder for the first time \cite{27}.

Ren \emph{et al.} theoretically studied the perception of vortex streets using real lateral line in 2010. Based on potential flow theory, they constructed the model of flow field around the fish, and then explained how the fish captured the characteristics of vortices with the help of lateral line CNs. The model were applied to estimate the range of the vortex, transmission speed, direction, distance between the vortex streets and distance between the fish and vortex street \cite{48}.

Klein \emph{et al.} used an artificial canal equipped with optical flow sensors which have been presented in section 2 (Figure \ref{Fig 8}(a)). They have demonstrated the capability of the ALL canals to detect the vibrating sphere. Additionally, vortices generated by an upstream cylinder were also detected. Based on the hydrodynamic information acquired by the ALL canal, they succeeded in calculating the flow velocity and the size of the cylinder \cite{24}.

In 2012, Venturelli \emph{et al.} used digital particle image velocimeter to visualize the flow state and a rigid body equipped with 20 pressure sensors (Figure \ref{Fig 10}(a)) parallel distributed to acquire flow field information and then applied time and frequency domain methods to describe hydrodynamic scenarios in steady and unsteady flows respectively. The array of pressure sensors showed a capability of discriminating vortex streets from steady flows and detecting the position and direction of the body relative to the incoming flow. A series of hydrodynamic parameters were also calculated, such as vortex shedding frequency, vortex travelling speed and downstream distance between vortices \cite{49}.

Free \emph{et al.} presented a method to estimate the parameters of vortices in 2017. They used a straight array of 4 pressure sensors for a spiral vortex and a square array for a K{\'a}rm{\'a}n vortex street. Based on potential flow theory and Bernoulli principle, the measurement equation was incorporated in a recursive Bayesian filter, as a consequence of which, the position and strength of vortices have been successfully estimated. Moreover, they identified an optimal path for underwater vehicles to swim through a K{\'a}rm{\'a}n vortex street using empirical observability. Experiments demonstrated the effectiveness of the closed-loop control \cite{50}. Based on the results above, in 2018, they installed the array on a Joukowski foil and detected K{\'a}rm{\'a}n vortex streets nearby. With the help of trajectory-tracking feedback control, the robotic foil performed fish-like slaloming behavior in a K{\'a}rm{\'a}n vortex street \cite{51}.

~\\
In this section, we have focused on the application of ALL system in detecting flow characteristics. Table 3 as follows lists different projects mentioned above and related ongoing studies. Existing results are mainly based on static ALL sensors and experiments are conducted in laboratory environment. The characteristics of flow which can be detected are also limited. For further study, with the improvement of ALL system, we can pay more attention to natural environment experiments, where the complexity of the water environment and the complex movements of the robot fish make it more difficult for perception.

\begin{longtable}[]{|c|c|c|c|}
	\caption{Classification of existing studies in hydrodynamic environment sensing and vortices detection}\\
	\hline
	Project                                                                                                  & Author                                                                                                    & ALL Sensors                                                                                                                                           & \begin{tabular}[c]{@{}c@{}}Laboratory experiment/ \\ Natural environment\\ experiment\end{tabular}     \\ \hline
	\multirow{3}{*}{\begin{tabular}[c]{@{}c@{}}Flow field \\ characteristics \\ identification\end{tabular}} & \begin{tabular}[c]{@{}c@{}}Kruusmaa et \\ al. 2013 \cite{73}\end{tabular}                                   & \begin{tabular}[c]{@{}c@{}}5 pressure sensors\\ (Intersema MS5407-AM)\end{tabular}                                                                    & Laboratory experiment                                                                                  \\ \cline{2-4}
	& \begin{tabular}[c]{@{}c@{}}Kruusmaa et \\ al. 2018 \cite{89}\end{tabular}                                   & \begin{tabular}[c]{@{}c@{}}11 pressure sensors\\ (SM5420C-030-A-P-S)\end{tabular}                                                                     & Laboratory experiment                                                                                  \\ \cline{2-4}
	& \begin{tabular}[c]{@{}c@{}}Liu et al. \\ 2019 \cite{52}\end{tabular}                                        & \begin{tabular}[c]{@{}c@{}}23 pressure sensors\\ (MS5803-07BA)\end{tabular}                                                                           & Laboratory experiment                                                                                  \\ \hline
	\multirow{4}{*}{\begin{tabular}[c]{@{}c@{}}Flow velocity \\ and direction \\ detection\end{tabular}}     & \begin{tabular}[c]{@{}c@{}}Kruusmaa et \\ al. 2013 \cite{73}\end{tabular}                                   & \begin{tabular}[c]{@{}c@{}}5 pressure sensors\\ (Intersema MS5407-AM)\end{tabular}                                                                    & Laboratory experiment                                                                                  \\ \cline{2-4}
	& \begin{tabular}[c]{@{}c@{}}Kruusmaa et \\ al. 2015 \cite{79}\\ Kruusmaa et \\ al. 2016 \cite{47}\end{tabular} & \multirow{2}{*}{\begin{tabular}[c]{@{}c@{}}16 pressure sensors\\ (SM5420C-030-A-P-S) \\ and 2 \\ three-axis accelerometers \\ (ADXL325BCPZ)\end{tabular}} & Laboratory experiment                                                                                  \\ \cline{2-2} \cline{4-4}
	& \begin{tabular}[c]{@{}c@{}}Kruusmaa et \\ al. 2016 \cite{86}\end{tabular}                                   &                                                                                                                                                       & \begin{tabular}[c]{@{}c@{}}Laboratory experiment \\ and natural environment \\ experiment\end{tabular} \\ \cline{2-4}
	& \begin{tabular}[c]{@{}c@{}}Liu et al.\\ 2020 \cite{56}\end{tabular}                                         & \begin{tabular}[c]{@{}c@{}}23 pressure sensors\\ (MS5803-07BA)\end{tabular}                                                                           & Laboratory experiment                                                                                  \\ \hline
	\multirow{6}{*}{\begin{tabular}[c]{@{}c@{}}Vortex street \\ properties \\ detection\end{tabular}}        & \begin{tabular}[c]{@{}c@{}}Yang et al.\\ 2006 \cite{27}\end{tabular}                                        & 16 HWA sensors                                                                                                                                        & Laboratory experiment                                                                                  \\ \cline{2-4}
	& \begin{tabular}[c]{@{}c@{}}Ren et al.\\ 2010 \cite{48}\end{tabular}                                         & \multicolumn{2}{c|}{Theoretical model}                                                                                                                                                                                                                         \\ \cline{2-4}
	& \begin{tabular}[c]{@{}c@{}}Klein et al.\\ 2011 \cite{24}\end{tabular}                                       & Optical sensors                                                                                                                                       & Laboratory experiment                                                                                  \\ \cline{2-4}
	& \begin{tabular}[c]{@{}c@{}}Venturelli et\\ al. 2012 \cite{49}\end{tabular}                                  & 20 pressure sensors                                                                                                                                   & Laboratory experiment                                                                                  \\ \cline{2-4}
	& \begin{tabular}[c]{@{}c@{}}Kruusmaa et \\ al. 2013 \cite{73}\end{tabular}                                   & \begin{tabular}[c]{@{}c@{}}5 pressure sensors\\ (Intersema MS5407-AM)\end{tabular}                                                                    & Laboratory experiment                                                                                  \\ \cline{2-4}
	& \begin{tabular}[c]{@{}c@{}}Free et al.\\ 2017 \cite{50}\\ Free et al.\\ 2018 \cite{51}\end{tabular}           & \begin{tabular}[c]{@{}c@{}}An array of 4 \\ pressure sensors\end{tabular}                                                                             & Laboratory experiment                                                                                  \\ \hline
\end{longtable}

\section{ALL based dipole source detection}
\begin{figure}[htbp]
	\centering
	\includegraphics[width=\columnwidth]{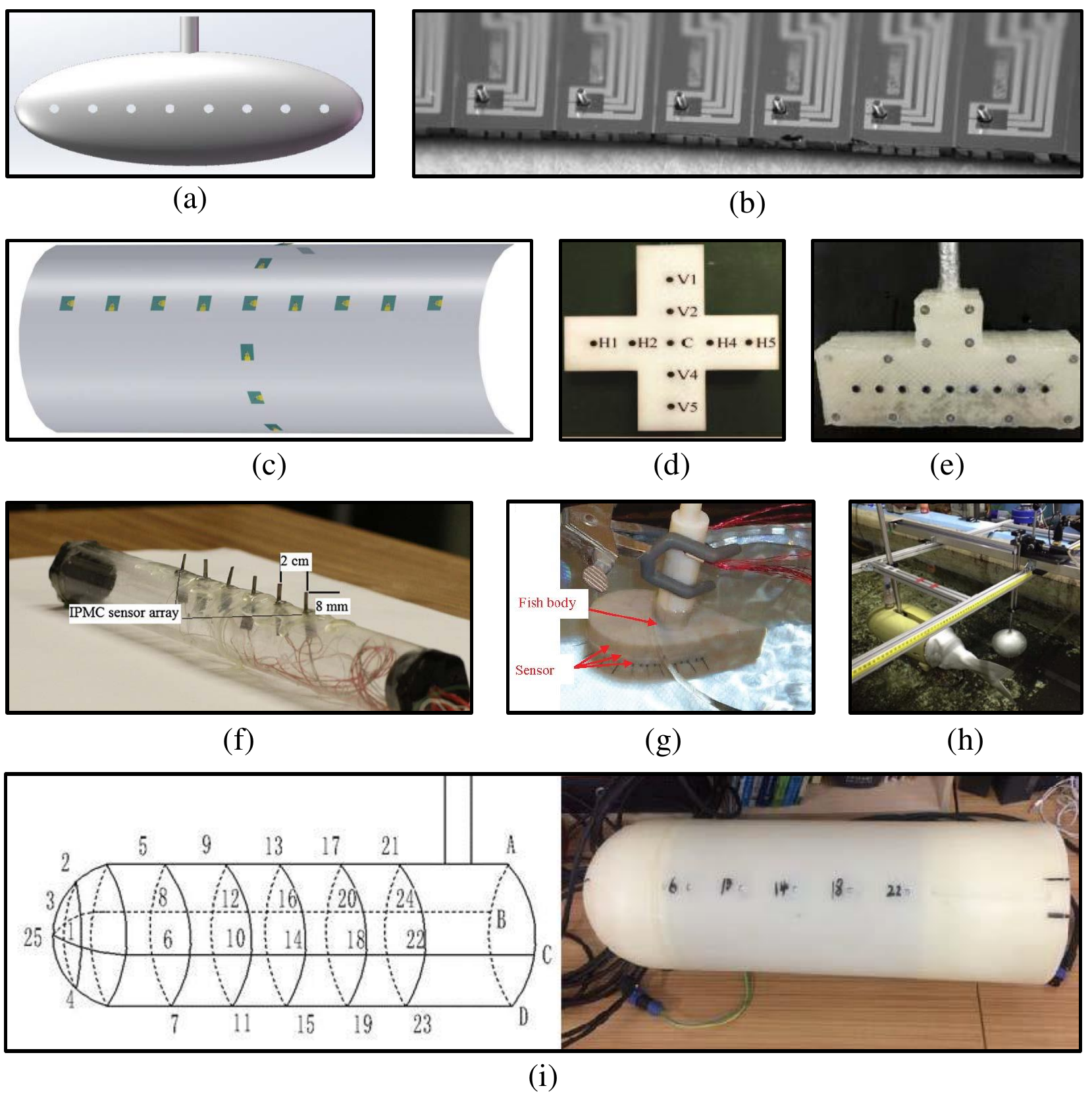}
	\caption{Different carriers boarded with ALL mentioned in Section 5. (a)The fish-shaped prototype inspired by the trout lateral line in \cite{90}. (b)Photo of an AHC sensor array used in \cite{4}. (c)Diagram of the ALL showing the biomimetic neuromast layout in \cite{6}. (d)The design of ALL in \cite{36}. (e)The design of ALL in \cite{37}. (f)An experimental prototype of IPMC-based lateral line system used in \cite{31}. (g)A prototype of an ALL with sensors used in \cite{35}. (h)Robotic fish with the PVDF sensor along with the oscillating sphere in \cite{41}. (i)Sensor layout and lateral line carrier physical map in \cite{40}.}
	\label{Fig 11}
\end{figure}

The localization ability of underwater objects with the help of ALL system can effectively improve the viability of robotic fish in underwater environment. In addition to the anti-K{\'a}rm{\'a}n vortex street, a near-dipole flow field is generated by the fin while fish swims. This can also explain how predators capture preys \cite{32}. Dipole oscillation source detection has become a common problem in hydrodynamics and the development of ALL. While the dipoles are vibrating or moving in a certain way, the pressure and the flow speed will change accordingly. By measuring these information with the help of ALL, we can infer the motion of the object for further study. Carriers boarded with ALL mentioned in this section are shown in Figure \ref{Fig 11}.

Tang \emph{et al.} represented an array of 8 pressure sensors installed on the surface of an underwater vehicle (Figure \ref{Fig 11}(a)) inspired by lateral line for near-field detection in 2019. The pressure field generated by vibrating sphere which was simulated as an underwater pressure source was derived by means of linearizing the kinematic and dynamic conditions of the free surface wave equation. The fish-shaped structure they used was boarded with 8 pressure sensors. The pressure field detected by the ALL was fit with simulation results \cite{90}.

In 2007, Yang \emph{et al.} used an array of AHCs (Figure \ref{Fig 11}(b)) whose sensory unit has been introduced in section 2 (Figure \ref{Fig 5}(b)) to locate and track the dipole source. As for mapping the pressure field produced by the dipole source, the array performed well and the results was consistent in experiments and theory. As for tracking the trail, a cylinder was put in a steady flow with the speed of 0.2 m/s to simulate the hydrodynamic trail which was dominated by K{\'a}rm{\'a}n vortex street consequently \cite{4}.

In 2010, the same group developed an ALL by wrapping 15 pressure sensor around a cylinder (Figure \ref{Fig 11}(c)) to mimic real fish and proved its localization capability. They used a beamforming algorithm to image hydrodynamic events in a 3-D domain. Consequently, the ALL sensors was demonstrated to be able to localize a dipole source and a tail-flicking crayfish accurately in varieties of conditions \cite{6}.

Additionally, Asadnia \emph{et al.} packaged Pb(Zr$_{0.52}$Ti$_{0.48}$)O$_3$ thin-film piezoelectric pressure sensors for underwater sensing in 2013. The array of 2 by 5 sensor has been showed in section 2 (Figure \ref{Fig 6}(a)). While the dipole was driven at the frequency of 15 Hz and moved parallel to the array, by measuring the maximum peak-to-peak output of the sensors, they approximately estimated the position of the dipole source. To detect the flow velocity generated by the dipole source, the array showed a resolution of 3 mm/s \cite{16}.

Abdulsadda \emph{et al.} put forward an array of ALL sensors based on the sensing capability of IPMC in 2011. The signals were processed through a widely-used neural network, which was similar to the biological counterpart. The ALL has been presented in the section 2 as well (Figure \ref{Fig 6}(c)). Experiments proved that the ALL could effectively locate the dipole source and the flapping tail. Moreover, the more sensors was used, the more precise the results were \cite{19}. In 2013, based on an analytical model of flow field produced by the dipole source, they presented another two schemes, Gauss Newton (GN) algorithms which were used to solve the nonlinear estimation problem by means of linear iteration and Newton Raphson (NR) algorithms which were used to solve the nonlinear equation under the condition of first-order optimality, to locate the dipole source and estimate amplitude and direction of the vibration. Additionally, they improved the design of intra-sensor spacing (Figure \ref{Fig 11}(f)) of the lateral line by analysis based on Cramer-Rao bound (CRB). With 19 dipole sources placed along an ellipsoidal track, the simulation and experiment results both proved the accuracy of this model \cite{31}\cite{33} \cite{34}. In order to reduce the influence of uncertainty in measurements and flow model and thus identify a vibrating dipole accurately, Tan \emph{et al.} developed a specialized bi-level optimization methodology to optimize the design parameters of the ALL \cite{35} (Figure \ref{Fig 11}(g)).

In the aspect of using hot-wire flow sensors to detect the dipole source, Pandya \emph{et al.} made great contributions in 2006. Section 2 has introduced the hot-wire ALL developed by them (Figure \ref{Fig 9}(a)). To make full use of the ALL, they reported on the implementation of a algorithm consisting of template training approach and the modeling approach based on minimum mean-squared error (MMSE) algorithm in order to locate and track a vibrational dipole source \cite{28}.

In 2010, by detecting the parallel and the perpendicular velocity components, Dagamseh \emph{et al.} used an array of hair flow sensors to reconstruct the velocity field induced by the dipole source in the air and measure the distance \cite{42} \cite{43} \cite{44} \cite{45}. In 2013, they employed beamforming techniques and improved the performance of the sensor array \cite{46}.

Inspired by the sensory abilities of lateral line, Zheng \emph{et al.} developed an ALL composed of 9 underwater pressure sensors forming a cross (Figure \ref{Fig 11}(d)) to locate a dipole source in 2018. For the sake of handling nonlinear pattern identification problem, they adopted the method of generalized regression neural network which performed well on condition that the array is below 13cm away from the dipole source \cite{36}. Besides, the same group develop another ALL composed of 9 pressure sensors in a straight line (Figure \ref{Fig 11}(e)) to locate the dipole source. Lin \emph{et al.} modelled the pressure field of the dipole source and obtained the position through least square method by means of the information acquired by the pressure sensors \cite{37}. Ji \emph{et al.} employed the same ALL and put forward a new method named MUSIC (multiple signal classification) for the purpose of locating dipole source with high-resolution based on spatial spectrum estimation. Moreover, they also presented a MVDR (minimum variance distortionless response) method which improved the previous Capon's method (an adaptive beamforming-based method). For further study, MUSIC method showed a potential to locate two close dipole sources \cite{38}. In 2019, Ji \emph{et al.} established a quantitive and method-independent Cramer-Rao lower bound (CRLB) model to evaluate the localization performance of the above two methods, least square and MUSIC, which provided guidance on the optimal design of the ALL with the least sensors and the most appropriate spacing \cite{39}.

In 2018, Liu \emph{et al.} developed an ALL consisting of 25 high precision pressure sensors (Figure \ref{Fig 11}(i)). They established a mathematical model of the dipole source through Euler equation and plane potential flow theory to describe the relationship between the characteristic parameters of source and the surface pressure of the underwater vehicle and successfully detected the position, frequency and amplitude of the source through a neural network model. The consistency of simulation and experiment results demonstrated the effectiveness of the method \cite{40}.

In 2018, Yen \emph{et al.} adopted the potential flow theory to predict the hydrodynamic pressure and presented a method to follow periodic stimuli generated by an oscillating source. The fin of the robot was regarded as an oscillator. By subtracting the pressure induced by the robotic fish from the pressure measured by the PVDF sensor (Figure \ref{Fig 11}(h)), they acquired the pressure generated by the source. Based on this, the robotic fish was able to adjust the amplitude, frequency, offset according to the phase difference \cite{41}. Furthermore, this method lays a solid foundation for controlling the robotic fish to swim in a school.

In 2019, Wolf \emph{et al.} used a 2-D array of 8 all-optical sensors to measure the velocity profiles of a underwater object and then adopted feed-forward neural network and recurrent neural network to reconstruct the position of the object \cite{57}\cite{58}. Furthermore, they implemented near field object classification based on hydrodynamic information with an Extreme Learning Machine neural network. This method provided more information about the shape comparing to other 2-D sensing array \cite{59}.

In this section, we have put emphasis on discussing applications of ALL in locating and tracking the dipole source, especially the difference in approaches to reaching the results. Table 4 as follows can be a summary of this section. Flow induced by oscillating sources is only one of the most basic forms of water flows and similar to wake flow generated by real fish. Based on the results above, we can also conduct natural environment experiments in which ALL is installed on robotic fish to follow real fish, even fish school and locate them in real time.

\begin{longtable}{|c|c|c|}
	\caption{Classification of existing studies in dipole source detection}\\
	\hline
	Author                                                                                                                                                   & ALL Sensors                                                                                      & Approaches                                                                                                                             \\ \hline
	\begin{tabular}[c]{@{}c@{}}Tang et al.\\ 2019 \cite{90}\end{tabular}                                                                                       & \begin{tabular}[c]{@{}c@{}}8 eight pressure\\ sensors\end{tabular}                               & \begin{tabular}[c]{@{}c@{}}Linearizing the kinematic\\ and dynamic conditions\\ of the free surface wave equation\end{tabular}         \\ \hline
	\begin{tabular}[c]{@{}c@{}}Yang et al.\\ 2007 \cite{4}\end{tabular}                                                                                        & \begin{tabular}[c]{@{}c@{}}An array of\\ AHC sensors\end{tabular}                                & \begin{tabular}[c]{@{}c@{}}Measuring the maximum\\ peak-to-peak out put\\ of the sensors\end{tabular}                                  \\ \hline
	\begin{tabular}[c]{@{}c@{}}Yang et al.\\ 2010 \cite{6}\end{tabular}                                                                                        & \begin{tabular}[c]{@{}c@{}}15 biomimetic\\ neuromasts\end{tabular}                               & Beamforming algorithm                                                                                                                  \\ \hline
	\begin{tabular}[c]{@{}c@{}}Pandya et \\ al. 2006 \cite{28}\end{tabular}                                                                                    & \begin{tabular}[c]{@{}c@{}}an array of 16\\ hot-wire anemometers\end{tabular}                    & \begin{tabular}[c]{@{}c@{}}Template training approach\\ and the modeling approach\end{tabular}                                         \\ \hline
	\begin{tabular}[c]{@{}c@{}}Asadnia et \\ al. 2013 \cite{16}\end{tabular}                                                                                   & \begin{tabular}[c]{@{}c@{}}An array of 2 by 5\\ pressure sensors\end{tabular}                    & Maximum pressure signal                                                                                                                \\ \hline
	\begin{tabular}[c]{@{}c@{}}Abdulsadda \\ et al. 2011 \cite{19}\end{tabular}                                                                                & 5 IPMC sensors                                                                                   & Neural network                                                                                                                         \\ \hline
	\begin{tabular}[c]{@{}c@{}}Abdulsadda et\\ al. 2013 \cite{31}\\ Abdulsadda et\\ al. 2012 \cite{33}\cite{34}\end{tabular}                                      & 6 IPMC sensors                                                                                   & \begin{tabular}[c]{@{}c@{}}Gauss Newton and\\ Newton Raphson algorithms\end{tabular}                                                   \\ \hline
	\begin{tabular}[c]{@{}c@{}}Ali Ahrari et \\ al. 2016 \cite{35}\end{tabular}                                                                                & \begin{tabular}[c]{@{}c@{}}Multiple flow\\ velocity sensors\end{tabular}                         & \begin{tabular}[c]{@{}c@{}}Bi-level optimization\\ methodology\end{tabular}                                                            \\ \hline
	\begin{tabular}[c]{@{}c@{}}Zheng et al.\\ 2018 \cite{36}\end{tabular}                                                                                      & \begin{tabular}[c]{@{}c@{}}9 underwater pressure\\ sensors forming a cross\end{tabular}          & \begin{tabular}[c]{@{}c@{}}Generalized regression\\ neural network\end{tabular}                                                        \\ \hline
	\begin{tabular}[c]{@{}c@{}}Lin et al.\\ 2018 \cite{37}\end{tabular}                                                                                        & \multirow{3}{*}{\begin{tabular}[c]{@{}c@{}}9 pressure sensors\\ in a straight line\end{tabular}} & Least square method                                                                                                                    \\ \cline{1-1} \cline{3-3}
	\begin{tabular}[c]{@{}c@{}}Ji et al.\\ 2018 \cite{38}\end{tabular}                                                                                         &                                                                                                  & MUSIC, MVDR                                                                                                                             \\ \cline{1-1} \cline{3-3}
	\begin{tabular}[c]{@{}c@{}}Ji et al.\\ 2019 \cite{39}\end{tabular}                                                                                         &                                                                                                  & CRLB model                                                                                                                             \\ \hline
	\begin{tabular}[c]{@{}c@{}}Liu et al.\\ 2018 \cite{40}\end{tabular}                                                                                        & \begin{tabular}[c]{@{}c@{}}25 high precision\\ pressure sensors\end{tabular}                     & \begin{tabular}[c]{@{}c@{}}Euler equation,\\ plane potential flow theory\\ and neural network\end{tabular}                             \\ \hline
	\begin{tabular}[c]{@{}c@{}}Yen et al.\\ 2018 \cite{41}\end{tabular}                                                                                        & a PVDF sensor                                                                                     & Potential flow theory                                                                                                                  \\ \hline
	\begin{tabular}[c]{@{}c@{}}Dagamseh et al.\\ 2009 \cite{42}\cite{43}\\ Dagamseh et al.\\ 2010 \cite{44}\cite{45}\\ Dagamseh et al.\\ 2013 \cite{46}\end{tabular} & \begin{tabular}[c]{@{}c@{}}an array of\\ hair flow-sensors\end{tabular}                          & \begin{tabular}[c]{@{}c@{}}Detecting the parallel\\ and the perpendicular\\ velocity components,\\ beamforming techniques\end{tabular} \\ \hline
	\begin{tabular}[c]{@{}c@{}}Wolf et al.\\ 2019\cite{57}\cite{58}\\Wolf et al.\\ 2020\cite{59}\end{tabular}                                                                               & \begin{tabular}[c]{@{}c@{}}8 all-optical\\ flow sensors\end{tabular}                             & \begin{tabular}[c]{@{}c@{}}Feed-forward neural network\\ and recurrent neural network\end{tabular}                                     \\ \hline
\end{longtable}

\section{Flow-aided control of underwater robots using ALL system}
\begin{figure}[htbp]
	\centering
	\includegraphics[width=\columnwidth]{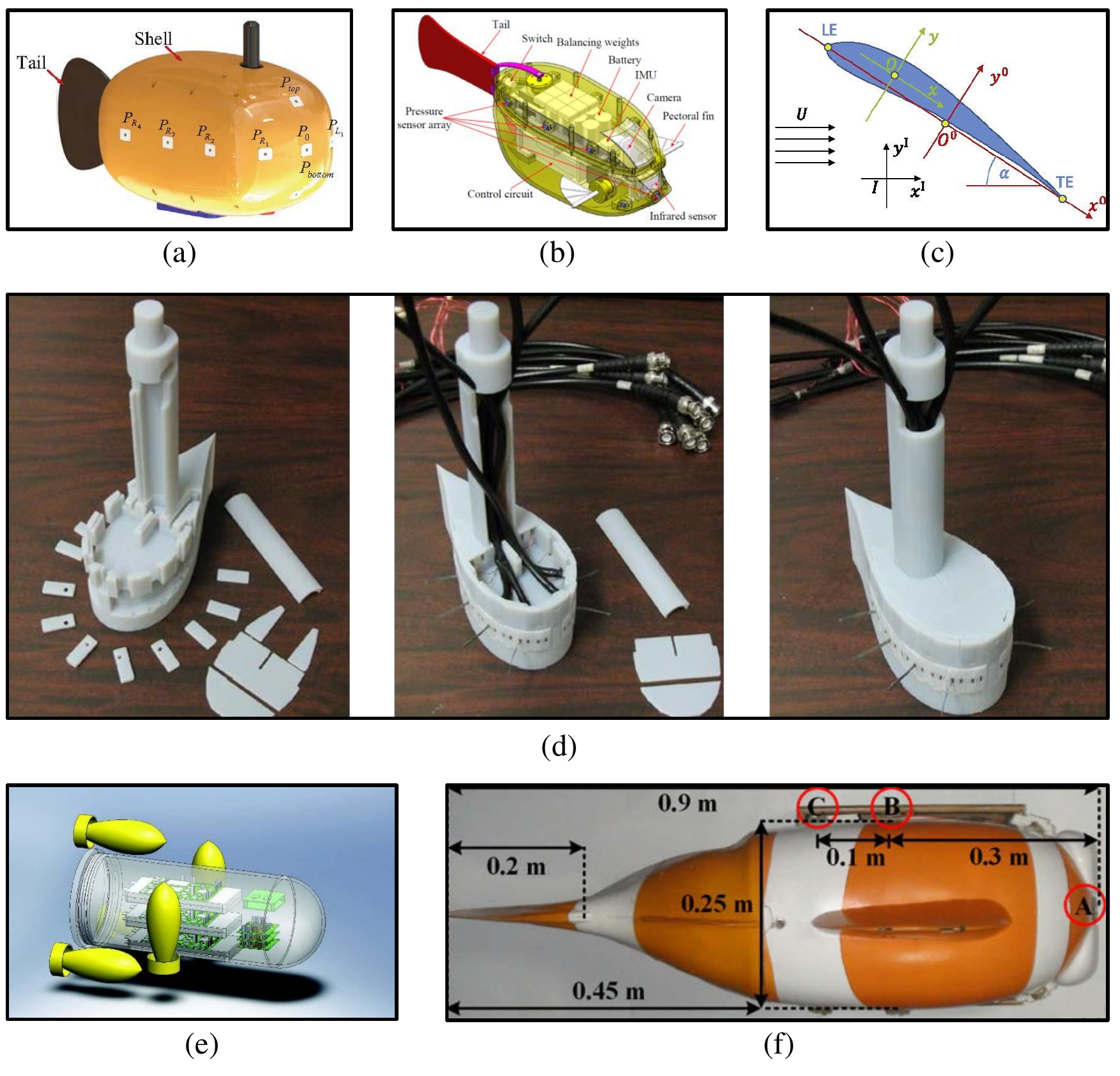}
	\caption{Different carriers boarded with ALL mentioned in this Section 6. (a)The robotic fish prototype in \cite{60}. (b)The mechanical structure and electronics of the robotic fish used in \cite{82}. (c)Illustration of reference frames I and O. TE denotes the trailing edge and LE the leading edge in \cite{85}. (d)Modular design of robotic foil in \cite{77}. An array of eight IPMC sensors are installed below an array of pressure sensors. (e)Conceptual 3-D drawing of the vehicle in \cite{84}. (f)Photographs of the robotic fish in \cite{93}.}
	\label{Fig 12}
\end{figure}

Lateral line plays an important part in sensing flow for fish school, which has inspired scientists worldwide to devote to developing underwater vehicles boarded with ALL system consisting of an array of sensors. The precious sections mainly presented applications based on a static ALL. If the ALL is moving like fish, the sensing difficulty will greatly increase. Assisted by ALL sensors, underwater vehicles are in a position to obtain fluid information accurately and effectively, providing the possibility to implement vehicles motion pattern identification and autonomous control. Varieties of experiments have been carried out in this domain, such as pattern identification, motion parameters (speed and direction) estimation and control, localization, obstacles detection and avoidance, energy consumption reduction and neighborhood robotic fish perception. Carriers boarded with ALL mentioned in this section are shown in Figure \ref{Fig 12}.

In 2013, Akanyeti \emph{et al.} firstly dived into the problem of hydrodynamic sensing on condition that the ALL is moving. Based on Bernoulli equation, they presented a formula about pressure detected by the ALL and the moving velocity and acceleration, which laid a solid foundation for the following study \cite{53}. Additionally, Chambers \emph{et al.} made a study of using a vertically or horizontally moving ALL to sense local flow. They came to a conclusion that a moving ALL performed better than static \cite{54}. For further studies, a moving carrier boarded with ALL which has a great sensing ability can provide more valuable experimental data. The project FILOSE using a robotic fish showed in Figure \ref{Fig 10}(b) not only aimed to study how fish perceive and respond to fluid simulation, but also constructed a bioinspired robot on the basis of it. Boarded with ALL sensors, robotic fish measured the data from surrounding fluid environment which provided information on hydrodynamic features. And then by analysis, the relationship between fluid data and kinestate of robotic fish was established, showing a new idea of robotic fish self-control. Several experiments have been conducted as follows: 1) detecting direction while swimming against the flow, 2) swimming along a predetermined trajectory, 3) maintaining stable position in constant current, 4) reducing energy consumption in turbulence, 5) reducing energy consumption by maintaining a stable position in the hydrodynamic shadow, 6) conducting control-experiments between real fish and robotic fish \cite{76}. In this section, we will introduce flow-aided control of underwater robots using ALL system by category.

\subsection{Pattern identification}
In 2020, Zheng \emph{et al.} made a breakthrough in motion parameters estimation of a robotic fish boarded with 11 pressure sensors (MS5803-01BA) (Figure \ref{Fig 12}(a)). When the robotic fish moved at a specific state, such as rectilinear motion, turning motion, gliding motion, and spiral motion, they established a model combining the motion parameter including linear velocity, angular velocity, motion radius, etc. and the superficial hydrodynamic pressure variations. Robotic fish acquired the motion parameters based on the pressure detected by the ALL and then predicted the trajectory \cite{62}. This work is of great importance for future study on self-trajectory-control.

In 2014, Liu \emph{et al.} conducted experiments that robotic fish (Figure \ref{Fig 12}(b)) sense pressure information while swimming in different gaits such as forward swimming, turning, ascending and diving. And then based on feature points extracted from the data, they adopted a subtractive clustering algorithm to recognize the swimming gaits of robotic fish \cite{88}. The success of this approach lays a solid foundation for quick control of robotic fish.

\subsection{Motion parameters (speed and direction) estimation and control}
In regard to the control of robotic fish, there have been a large number of results as well. Instantly, Kruusmaa \emph{et al.} implemented rheotaxis behaviour in robotic fish in 2011. With the pressure sensors detecting flow information, they put forward a linear control law which helped the robotic fish to adjust the beat frequency in order to maintain position in the steady flow \cite{55}. In 2013, they used a 50cm-long robotic fish (Figure \ref{Fig 10}(b)) mimicking the geometry and swimming mode of a rainbow trout and put forward a formula for estimating the speed. The experiments have proved the validity of it \cite{73}. Inspired by the Braitenberg vehicle 2b, they installed two pressure sensors on both sides of the head which could detect the pressure difference on the left and right side of the robotic fish \cite{81}. And then it was able to change the direction of swimming according to the difference to remain stable. Ulteriorly, they realized the position estimation and position stability of the robotic fish in steady water flows and behind solid objects \cite{73}.

Additionally, Xie \emph{et al.} designed a robotic fish inspired by the geometry and swimming pattern of an ostraciiform boxfish shown in Figure \ref{Fig 12}(b). The robotic fish is boarded with an ALL composed of an array of 11 pressure sensors (Consensic CPS131) and an inertial measurement unit (IMU). The former is used for fluid dynamic pressure data acquisition while the latter is used to monitor the robot pitch, yaw and roll angles. In order to reduce errors on account of sensors' inaccuracy and instability, they employed an optimal information fusion decentralized filter in 2015, as a consequence of which, the accuracy of speed estimation has been greatly improved. The speed estimation formula is derived from Bernoulli principle and corrected by local filter from ALL and IMU \cite{82}. Furthermore, in 2016, they put forward a nonlinear prediction model including distributed pressure and angular velocity to estimate the speed of robotic fish \cite{75}.

Moreover, Paley \emph{et al.} also presented something new in flow speed and angle-of-attack detecting in 2015. They used a new type of flexible robotic fish whose shape is Joukowski-airfoil (Figure \ref{Fig 12}(c)) with distributed pressure sensors. Flow speed, angle-of-attack, and foil surfaces were estimated by a recursive Bayesian filter assimilation pressure measurement. And they combined an inverse-mapping feedforward controller based on an average model derived for periodic actuation of angle-of-attack and a proportional-integral feedback controller utilizing the estimated flow information to implement the closed-loop speed-control strategy \cite{77}\cite{85}\cite{92}.

\subsection{Obstacles detection and avoidance}
Furthermore, extensive efforts have been done in obstacle detection and avoidance. Inspired by sensitivity of lateral line to the presence of oncoming currents and walls or obstacles, Paley \emph{et al.} developed a kind of wing underwater vehicle boarded with ALL sensors in 2015 (Figure \ref{Fig 12}(d)). Employing potential flow theory, they simulated the flow field around vehicle in the situation that the flow was uniform and there were obstacle upstream. A nonlinear estimation model of free stream flow speed, attack of angle and relative position of obstacles by measuring local flow speed and pressure difference was derived theoretically. In order to implement the stability of swimming direction and position behind obstacles, they presented a recursive Bayesian filter. Finally, they discussed the Closed-loop control strategy ulteriorly \cite{77}.

In addition, Martiny \emph{et al.} studied intensively in obstacles detection and avoidance in 2009. They developed an autonomous underwater vehicle equipped with 4 ALL sensors (Figure \ref{Fig 12}(e)). Using hot-wire anemometry, it measured local flow speed around the vehicle, which was proved related to the distance between obstacles and the vehicle theoretically and experimentally \cite{84}.

Yen \emph{et al.} have made a breakthrough in obstacles detection and navigation in 2017. They used a robotic fish boarded with 3 ALL sensors (Figure \ref{Fig 12}(f)) to measure nearby pressure variations, on the basis of which, they presented a way to control a robotic fish to swim along a straight wall. In theory, the tail of the robotic fish was regarded as an oscillating dipole in a 2-D potential flow approximately and the wall effect was described by an image dipole on the opposite side of the wall. The robotic fish responded to the pressure variations in order to keep a fixed distance from the wall. A qualitative relationship between velocity and wall effect was concluded in this research \cite{93}.

\subsection{Neighborhood robotic fish perception}
Not only have there been various results with respect to a single robotic fish, large amounts of experiments have also been carried out in multi-body control. Lots of work in this research field has been done by Xie \emph{et al.} In 2015, they came to the conclusion by experiments that robotic fish (Figure \ref{Fig 12}(b)) was capable of sensing the beating frequency of the robot swimming in front and the distance between the two robots with the help of ALL system \cite{74}. In 2017, they used the ALL system to detect the reverse K{\'a}rm{\'a}n-vortex street-like vortex wake generated by its adjacent robotic fish. By extracting meaningful information from the pressure variations caused by the reverse K{\'a}rm{\'a}n-vortex street-like vortex wake, the oscillating frequency/amplitude/offset of the adjacent robotic fish, the relative vertical distance and the relative yaw/pitch/roll angle between the robotic fish and its neighbor were sensed efficiently \cite{83}. This progress lays a solid foundation for multi-body interactions research in the future.

In 2019, Zheng \emph{et al.} used a robotic fish which is shown in Figure \ref{Fig 12}(a) to conduct neighborhood perception experiments. Firstly, based on Bernoulli principles, they established a theoretical model to describe the hydrodynamic pressure variations on the surface of two adjacent robotic fish which swam diagonally ahead another \cite{60}. Then, they utilized dye injection technique, hydrogen bubble technique and computational fluid dynamics simulation to study the vortices induced by the robotic fish separation. Besides, on the basis of the previous model, they also presented the relationship describing longitudinal separations and superficial hydrodynamic pressure variations of two robotic fish \cite{61}. This progress provided a new method for robotic fish school sensing.

~\\n addition to the results mentioned above, some other applications of ALL in robotic fish control are introduced below. With respect to localization, Muhammad \emph{et al.} produced preliminary results by flow feature extraction and comparison of compact flow features, based on which they developed an underwater landmark recognition technique in 2015 \cite{80}. This technique enables robotic fish to recognize locations that it has previously visited both in semi-natural and natural environments. In 2017, Francisco \emph{et al.} proposed a map-based localization technique that employd simulated hydrodynamic maps. They used a computational fluid dynamics model to generate a flow rate diagram. Hydrodynamic information was acquired by ALL systems and analyzed to estimate the speed. Compared with the flow rate diagram, the location of the robotic fish was found out in the flow rate maps which was simulated from hydrodynamic results \cite{87}.

As for energy consumption, Kruusmaa \emph{et al.} conducted a control experiment that the robotic fish swims in steady flow, behind a cylinder and behind a cuboid in 2013. Consequently, swimming behind a cuboid consumed the least energy. They assumed that this phenomenon was on account of the presence of the well-defined suction zone behind the cylinder. Both obstacles avoidance and energy consumption reduction are of giant significance to the navigation of robotic fish \cite{73}.

In this section, we have focused on the application of ALL system in robotic fish control. Table 5 as follows lists different projects mentioned above and related ongoing studies. Similar to the previous sections, robotic fish in this section is mostly static or move in a simple state, such as rectilinear motion and turning motion in laboratory environment. However, the motion of real fish and the real underwater environment are much more complicated, which makes it more difficult for underwater sensing. To solve these problems, we need to improve the sensing system and establish new control algorithms for natural environment studies. Additionally, perception of underwater obstacles provides a new approach for underwater environment reconstruction and results of neighborhood robotic fish perception can be a basis of control of multi robotic fish, both of which are potential to promote underwater exploration.

\begin{longtable}[]{|c|c|c|c|}
	\caption{Classification of existing studies in flow-aided control}\\
	\hline
	Project                                                                                  & Author                                                                                                      & ALL Sensors                                                                                                                                                                                                                           & \begin{tabular}[c]{@{}c@{}}Laboratory experiment/\\ Natural environment\\ experiment\end{tabular} \\ \hline
	\multirow{2}{*}{\begin{tabular}[c]{@{}c@{}}Pattern\\ identification\end{tabular}}        & \begin{tabular}[c]{@{}c@{}}Liu el al.\\ 2014 \cite{88}\end{tabular}                                           & \begin{tabular}[c]{@{}c@{}}9 pressure \\ sensors (CPS131)\end{tabular}                                                                                                                                                                & \begin{tabular}[c]{@{}c@{}}Laboratory \\ experiment\end{tabular}                                  \\ \cline{2-4}
	& \begin{tabular}[c]{@{}c@{}}Zheng et al.\\ 2020 \cite{62}\end{tabular}                                         & \begin{tabular}[c]{@{}c@{}}11 pressure sensors\\ (MS5803-14BA)\end{tabular}                                                                                                                                                           & \begin{tabular}[c]{@{}c@{}}Laboratory\\ experiment\end{tabular}                                   \\ \hline
	\multirow{2}{*}{\begin{tabular}[c]{@{}c@{}}Direction\\detection and\\holding\end{tabular}}   & \begin{tabular}[c]{@{}c@{}}Kruusmaa et \\ al. 2012 \cite{81}\\ Kruusmaa et \\ al. 2013 \cite{73}\end{tabular}   & \begin{tabular}[c]{@{}c@{}}5 pressure sensors\\ (Intersema MS5407-AM)\end{tabular}                                                                                                                                                    & \begin{tabular}[c]{@{}c@{}}Laboratory \\ experiment\end{tabular}                                  \\ \cline{2-4}
	& \begin{tabular}[c]{@{}c@{}}Paley et al. \\ 2013 \cite{92}\\ Paley et al. \\ 2015 \cite{77}\cite{85}\end{tabular} & \begin{tabular}[c]{@{}c@{}}8 IPMC sensors and \\ four embedded \\ pressure sensors \cite{77}/\\ six pressure sensors \\ (Servoflo MS5401-BM) \cite{85}/\\ The sensors (MikroTip \\ Catheter Pressure \\ Transducers) \cite{92}\end{tabular} & \begin{tabular}[c]{@{}c@{}}Laboratory\\ experiment\end{tabular}                                   \\ \hline
	\multirow{3}{*}{\begin{tabular}[c]{@{}c@{}}Speed\\ estimation\end{tabular}}              & \begin{tabular}[c]{@{}c@{}}Kruusmaa et \\ al. 2013 \cite{73}\end{tabular}                                     & \begin{tabular}[c]{@{}c@{}}5 pressure sensors\\ (Intersema MS5407-AM)\end{tabular}                                                                                                                                                    & \begin{tabular}[c]{@{}c@{}}Laboratory\\ experiment\end{tabular}                                   \\ \cline{2-4}
	& \begin{tabular}[c]{@{}c@{}}Xie et al.\\ 2015 \cite{82}\\ Xie et al.\\ 2016 \cite{75}\end{tabular}               & \begin{tabular}[c]{@{}c@{}}11 pressure sensors\\ (Consensic CPS131)\end{tabular}                                                                                                                                                      & \begin{tabular}[c]{@{}c@{}}Laboratory\\ experiment\end{tabular}                                   \\ \cline{2-4}
	& \begin{tabular}[c]{@{}c@{}}Paley et al.\\ 2013 \cite{92}\\ Paley et al.\\ 2015 \cite{77} \cite{85}\end{tabular}   & \begin{tabular}[c]{@{}c@{}}8 IPMC sensors and \\ four embedded \\ pressure sensors \cite{77}/\\ six pressure sensors \\ (Servoflo MS5401-BM) \cite{85}/\\ The sensors (MikroTip \\ Catheter Pressure \\ Transducers) \cite{92} \end{tabular} & \begin{tabular}[c]{@{}c@{}}Laboratory\\ experiment\end{tabular}                                   \\ \hline
	\begin{tabular}[c]{@{}c@{}}Position\\ holding\end{tabular}                               & \begin{tabular}[c]{@{}c@{}}Kruusmaa et \\ al. 2013 \cite{73}\end{tabular}                                     & \begin{tabular}[c]{@{}c@{}}5 pressure sensors\\ (Intersema MS5407-AM)\end{tabular}                                                                                                                                                    & \begin{tabular}[c]{@{}c@{}}Laboratory \\ experiment\end{tabular}                                  \\ \hline
	Localization                                                                             & \begin{tabular}[c]{@{}c@{}}Kruusmaa et \\ al. 2015 \cite{80}\\ Kruusmaa et \\ al. 2017 \cite{87}\end{tabular}   & \begin{tabular}[c]{@{}c@{}}14 pressure sensors\\ (Intersema MS5407-AM) \cite{80}/\\ 16 pressure sensors \cite{87}\end{tabular}                                                                                                            & \begin{tabular}[c]{@{}c@{}}Natural environment \\ experiment\end{tabular}                         \\ \hline
	\multirow{3}{*}{\begin{tabular}[c]{@{}c@{}}Obstacles\\detection and\\avoidance\end{tabular}} & \begin{tabular}[c]{@{}c@{}}Paley et al.\\ 2015 \cite{77}\end{tabular}                                         & \begin{tabular}[c]{@{}c@{}}8 IPMC sensors and \\ four embedded pressure \\ sensors\end{tabular}                                                                                                                                       & \begin{tabular}[c]{@{}c@{}}Laboratory\\ experiment\end{tabular}                                   \\ \cline{2-4}
	& \begin{tabular}[c]{@{}c@{}}Martiny et \\ al. 2009 \cite{84}\end{tabular}                                      & 4 pressure sensors                                                                                                                                                                                                                    & \begin{tabular}[c]{@{}c@{}}Natural environment\\ experiment\end{tabular}                          \\ \cline{2-4}
	& \begin{tabular}[c]{@{}c@{}}Yen et al.\\ 2018 \cite{93}\end{tabular}                                           & \begin{tabular}[c]{@{}c@{}}3 pressure sensors\\ (MS5803-01BA)\end{tabular}                                                                                                                                                            & \begin{tabular}[c]{@{}c@{}}Laboratory \\ experiment\end{tabular}                                  \\ \hline
	\begin{tabular}[c]{@{}c@{}}Energy\\consumption\\reducement\end{tabular}                  & \begin{tabular}[c]{@{}c@{}}Kruusmaa et \\ al. 2014 \cite{73}\end{tabular}                                    & \begin{tabular}[c]{@{}c@{}}5 pressure sensors\\ (Intersema MS5407-AM)\end{tabular}                                                                                                                                                    & \begin{tabular}[c]{@{}c@{}}Laboratory \\ experiment\end{tabular}                                  \\ \hline
	\multirow{2}{*}{\begin{tabular}[c]{@{}c@{}}Neighboring \\ robot fish \\ sensing\end{tabular}}    & \begin{tabular}[c]{@{}c@{}}Xie et al.\\ 2015 \cite{74}\\ Xie et al.\\ 2017 \cite{83}\end{tabular}               & \begin{tabular}[c]{@{}c@{}}9 pressure sensors\\ (MS5803-01BA)\end{tabular}                                                                                                                                                            & \begin{tabular}[c]{@{}c@{}}Laboratory \\ experiment\end{tabular}                                  \\ \cline{2-4}
	& \begin{tabular}[c]{@{}c@{}}Zheng et al.\\ 2019 \cite{60} \cite{61}\end{tabular}                                 & \begin{tabular}[c]{@{}c@{}}11 pressure sensors\\ (MS5803-01BA)\end{tabular}                                                                                                                                                           & \begin{tabular}[c]{@{}c@{}}Laboratory \\ experiment\end{tabular}                                  \\ \hline
\end{longtable}

\section{Discussion}
In the previous sections, we have reviewed ALL sensors based on different principles and applications in flow field characteristics identification, dipole source detection and control of underwater robots. Although fish lateral line provides inspiration for the design of ALL sensors and underwater detection, it also serves as a strict standard for research. There have been great progress in this area, however, performance of existing ALL system is still quite far from that of real fish.

ALL sensors which have been developed are no match for that of real fish through evolution in sensitivity, stability, coordination and information processing. We can optimize the design of the sensitive element and consider the resonance frequency for a better perception of ALL sensors. As for the stability, measurement errors in different temperature or pressure conditions should be taken into account. Additionally, waterproofing measures is necessary for the normal operation of sensors in the harsh conditions underwater. The development of new materials and micromachining technology provides possible methods for improvements in both areas. Not only should the performance of a single sensory unit be improved, the coordination of an array of ALL sensors is also important. Existing arrays of ALL sensors are mainly composed of a single type of sensors (pressure sensors or flow sensors) and arranged regularly, which is quite from that of real fish. Real lateral line consists of SNs and CNs for a comprehensive perception of surrounding environment and sensing cells are distributed in a specific pattern for a better sensing. Using pressure sensors and flow sensors simultaneously is a potential method to optimize the array of ALL sensors. Besides, we can put forward evaluation indexes of ALL in order to find out the best placement of sensors on the surface of underwater robots. In terms of information processing, the fish lateral line has many different sensory functions, which can be a reference standard for ALL sensors. Many algorithms in velocity measurements and dipole source detection have been put forward, but the sensory ability of ALL does not stop here. ALL has the potential to sense the obstacles, fish schools and even reconstruct the surrounding water environment, which is a basis of follow-up research on control of underwater robots.

As for hydrodynamic characteristics identification, there have been many results based on ALL system. However, existing results are mainly based on laboratory experiments where the motion of ALL carrier is simple such as static state and rectilinear motion and the water environment is stable. By contrast, the motion of real fish and underwater environment are much more complicated, which will greatly increase the difficulty of experiments. With the development of ALL sensors, natural experiments are necessary for complete imitation of real fish lateral line.

In dipole source detection, many methods and algorithms have been presented because oscillatory flow is one of the most basic flows and can be used to simulate the wake produced by the wagging tail of fish, which is important for further study on location and track of real fish. But the matching degree of oscillating flow and fish wake needs further exploration. On the basis of it, more experiments on real-time location of fish school and mimicking the group behavior of fish may become a focus in the future.

Flow-aided control of underwater robots is a promising project which has a wide range of applications in marine exploration. The first priority is to establish the motion model of underwater robots based on hydrodynamics theory or date-driven methods which provides a way for self-identification of motion pattern in the conditions where it is impossible to observe. Establishment of model can be a theoretical instruction for control strategies. In addition, obstacles recognition and avoidance is another necessary ability for autonomous control. Based on the location of obstacles, robots is potential to reconstruct the surrounding environment and plan an optimal path for navigation. Except for a single robotic fish, perception and control of a robotic fish group may greatly improve the efficiency of underwater exploration and the success rate of underwater missions with the help of ALL.

\section{Conclusion}
In this article, we briefly introduced the morphology and mechanism of the lateral line for a further understanding. And then we put emphasis on discussing progress in biomimetics inspired by lateral line. Different kinds of sensors based on different principles have been explored, which provides a new approach for fluid information detection. Furthermore, scientists have arranged sensors to fabricate ALL system on underwater vehicle bodies to assist in underwater detection, which is superior to traditional methods in applicability and flexibility. For hydrodynamic environment sensing and characteristic detection, we have made great progress in flow field characteristics identification, flow velocity and direction detection, vortex street properties detection. Additionally, various algorithms have been put forward to locate the dipole source. Recently, ALL system have been used for the autonomous control of underwater robotic fish.

In future, based on existing results, we need to develop sensor units demonstrating higher sensitivity and optimize the distribution of ALL on the robotic fish surface. Combining a high-performance sensor array with a more efficient signal processing algorithm, robotic fish is potential to show a sensory ability close to real fish, which is a foundation for natural environment experiments in the following studies. It is necessary to present new universal control algorithms for the autonomous control of robotic fish based on ALL systems. Perception of underwater obstacles provides a new approach for underwater environment reconstruction, which is important for underwater location and navigation. Moreover, application of ALL in the control of robotic fish school will also become a focus issue, which is important in multi-robot interaction and cooperation for future ocean exploration. With the development of new materials, fabrication technology and artificial intelligence, the performance of ALL system will be comparable to that of real fish.

The development of ALL systems provides a powerful tool for ocean exploration. Although great efforts have been devoted in this area, there is a long way to go to realize the comparable sensory ability and autonomous control of robotic fish to real fish.

\ack
This work was supported in part by grants from the National Natural Science Foundation of China (NSFC, No. 91648120, 61633002, 51575005) and the Beijing Natural Science Foundation (No. 4192026).\\

\section*{Reference}

\bibliography{ref}

\end{document}